\documentclass[fleqn,10pt]{wlscirep}
\usepackage[utf8]{inputenc}
\usepackage[T1]{fontenc}
\usepackage{color, soul}
\title{Self Reward Design with Fine-grained Interpretability}

\author[1,2,*]{Erico Tjoa}
\author[1]{Cuntai Guan}
\affil[1]{Nanyang Technological University, Singapore}
\affil[2]{Alibaba Group}
\affil[*]{Corresponding author}


\begin{abstract}
The black-box nature of deep neural networks (DNN) has brought to attention the issues of transparency and fairness. Deep Reinforcement Learning (Deep RL or DRL), which uses DNN to learn its policy, value functions etc, is thus also subject to similar concerns. This paper proposes a way to circumvent the issues through the bottom-up design of neural networks with detailed interpretability, where each neuron or layer has its own meaning and utility that corresponds to humanly understandable concept. The framework introduced in this paper is called the Self Reward Design (SRD), inspired by the Inverse Reward Design, and this interpretable design can (1) solve the problem by pure design (although imperfectly) and (2) be optimized like a standard DNN. With deliberate human designs, we show that some RL problems such as lavaland and MuJoCo can be solved using a model constructed with standard NN components with few parameters. Furthermore, with our fish sale auction example, we demonstrate how SRD is used to address situations that will not make sense if black-box models are used, where humanly-understandable semantic-based decision is required.
\end{abstract}
\begin{document}

\flushbottom
\maketitle
\thispagestyle{empty}

\section*{Introduction}
Reinforcement Learning (RL) and Deep Neural Network (DNN) have recently been integrated into what is known as the Deep Reinforcement Learning (DRL) to solve various problems with remarkable performance. DRL greatly improves the state-of-the-art of control and, in the words of Sutton and Barto\cite{1043175824}, \textit{learning from interaction}. Among the well-known successes are (1) the Deep Q-Network\cite{Mnih2015} which enabled machine to play Atari Games with incredible performance, and (2) AlphaGo that is capable of playing notoriously complex game of Go\cite{Silver2016} at and beyond pro human level. Although DNN has proven to possess great potentials, it is a black-box that is difficult to interpret. To address this difficulty, various works have emerged, thus we have a host of different approaches to eXplainable Artificial Intelligence (XAI); see surveys \cite{BARREDOARRIETA202082, 8631448, 9233366}. They have shed some lights into the inner working of a DNN, but there may still be large gaps to fill. Note that there is no guarantee that interpretability is even attainable, especially when context-dependent interpretability can be subjective. 

In this paper, we propose the Self Reward Design (SRD), a non-traditional RL solution that combines highly interpretable human-centric design and the power of DNN. Our robot (or agent, interchangeably used) rewards itself through purposeful design of DNN architecture, enabling it to partially solve the problem without training. While the initial hand-designed solution might be sub-optimal, the use of trainable DNN modules allows it to be optimized. We show that deep-learning style training might improve the performance of an SRD model or alter the system's dynamic in general. 

This paper is arranged as the following. We start with clarifications. Then we briefly go through related works that inspire this paper. Then our interpretable design and SRD optimization are demonstrated with a 1D toy example, RobotFish. We then extend our application to the Fish Sale Auction scenario which we go through more extensively in the main text. It is then followed by brief descriptions of how SRD can be used in 2D robot in the lavaland and MuJoCo simulation (their details in the appendix). We largely focus on the interpretability of design and SRD training, although we also include concepts like unknown avoidance, imagination and continuous control. All codes are available in \url{https://github.com/ericotjo001/srd} where the link to our full data can be found.

\section*{This Paper Focuses Heavily on Interpretable Human Design}
\label{section:focus}
\textbf{What exactly is this paper about?} This paper comprises demonstrations of how some reinforcement learning (RL) problems can be solved in an interpretable manner through self-reward mechanism. Our fish sale auction example also demonstrates an RL-like system that requires decision-making in RL style, but requires high human interpretability and is thus not fully compatible with standard RL optimization. Readers will be introduced to the design of different components tailored to different parts of the problems in a humanly understandable way.

 \textit{Important note: significance and caveat}. The paper has been reorganized to direct readers' focus to \textit{interpretable design} since reviewers tend to focus on standard RL practice instead of focusing on our main proposal, which is the interpretable design. This paper demonstrates the integration of a system that \underline{heavily uses human design augmented by NN}. Through this paper, we hope to encourage deep learning practitioners to develop transparent, highly interpretable NN-based reinforcement learning solutions in contrast to standard DRL models with large black-box components. Our designs can be presented in a way that are meaningful down to the granular level. \underline{What we do not claim}: we do NOT claim to achieve any extraordinary performance, although our systems are capable of solving the given problems. 

\textbf{But what is interpretability?} While there may be many ways to talk about interpretability, interpretability in the context of this paper is fine-grained, i.e. we go all the way down to \textit{directly manipulating weights and biases} of DNN modules. DNN modules are usually optimized using gradient descent from random initialization, thus the resulting weights are hard to interpret. In our SRD model, the meaning and purpose of each neural network component can be explicitly stated with respect to the environmental and model settings. 

\textit{How do we compare our interpretability} with existing explainable deep RL methods? Since we directly manipulate the weights and biases, our interpretability is at a very low level of abstraction, unlike post-hoc analysis e.g. saliency\cite{pmlr-v80-greydanus18a} or semantically meaningful high level specification such as reward decomposition\cite{650899}. In other words, we aim to be the most transparent and interpretable system, allowing users to understand the model all the way down to its most basic unit. Unfortunately, this means numerical comparison of interpretability does not quite make sense.

\textbf{Baseline}. We believe comparing performance with other RL methods is difficult since powerful DRL is likely to solve some problems very well w.r.t some measure of accuracy. Furthermore, not only are they often black-boxes that do not work in humanly-comprehensible way, sometimes their reproducibility is not very straightforward\cite{henderson2018deep}. Most importantly, in the context of this paper, focusing on performance distracts readers from our focus on interpretability. If possible, \textit{we want to compare the level of interpretability}. However, quantitative comparison is tricky, and we are not aware of any meaningful way to quantify interpretability that can compare with our proposed fine-grained model. So, what baseline should be used? The short answer is: there is no baseline to measure our type of interpretability. As previously mentioned, this is because our interpretability is fine-grained as we directly manipulate weights and biases. In this sense, our proposed methods are already the most interpretable system, since each basic unit has a specific, humanly understandable meaning. Furthermore, the set up of our auction experiments is not compatible with the concept of reward maximization used in standard RL, rendering comparison of ``quality" less viable. In any case, our 2D robot lavaland example achieves a high performance of approximately 90\% accuracy, given 10\% randomness to allow for exploration, which we believe is reasonable.

\subsection*{Related Works: From RL to Deep RL to SRD}
\label{section:relatedworks}
\textbf{RL with imperfect human design}. RL system can be set up by human manually specifying the rewards. Unfortunately, human design can easily be imperfect since the designers might not necessarily grasp the full extent of complex problem. For RL agents, designers' manual specification of rewards are fallible, subject to errors and problems such as \textit{negative side effect of a misspecified reward}\cite{FaultyReward}  and \textit{reward hacking}\cite{russel2010}. Dylan's \textit{inverse reward design} (IRD) paper\cite{10.5555/3295222.3295421} addresses this problem directly: it allows a model to learn beyond what imperfect designers specify - also see the appendix regarding Reward Design Problem (RDP). In this paper, the initialization of our models is generally also imperfect although we use SRD to meaningfully optimize the parameters. Another example of our solution to the imperfect designer problem is through \textit{unknown avoidance}, particularly \(w_{unknown}\) in lavaland problem. 

\textbf{From RL to Deep RL to interpretable DRL}. Not only is human design fallible, a good design may be difficult to create especially for complex problems. In the introduction, we mention that DRL solves this by combining RL and the power of DNN. However, DRL is a black-box that is difficult to understand. Thus the study of explainable DRL emerges; RL papers that address explainability/interptretability problems have been compiled in some survey papers\cite{HEUILLET2021106685,10.1007/978-3-030-57321-8_5}. Saliency, a common XAI method, has been applied to visualize deep RL's mechanism\cite{pmlr-v80-greydanus18a}. Relational deep RL uses a relational module that not only improves the agent's performance on StarCraft II and Box-World, but also provides visualization on the attention heads useful for interpretability\cite{zambaldi2018deep}. Other methods to improve interpretability include reward decomposition, in which each part of the decomposable reward is semantically meaningful\cite{650899}; do refer to the survey papers for several other ingenious designs and investigations into the interpretability of deep RL. 

In particular, explainable DRL models with manually specified humanly understandable tasks have emerge. Programmatically Interpretable RL\cite{verma2018programmatically} (PIRL) is designed to find programmatic policies for semantically meaningful tasks, such as car acceleration or steering. Symbolic techniques can further be used for the verification of humanly interpretable logical statements. Automated search is performed with the help of an oracle i.e. the interpretable model ``imitates" the oracle, eventually resulting in an interpretable model with comparable performance. Their interpretable policies consist of clearly delineated logical statements with some linear combinations of operators and terms. The components compounded through automatic searches might yield convoluted policies, possibly resulting in some loss of interpretability. By contrast, our model achieves interpretability by the strengths of activation of semantically meaningful neurons which should maintain their semantic assuming no extreme modification is performed. 

Multi-task RL\cite{shu2017hierarchical} uses \textit{hierarchical policy architecture} so that its agent is able to select a sequence of humanly interpretable tasks, such as ``get x" or ``stack x", each with its own policy \(\pi_k\). Each sub-policy can be separately trained on a sub-scenario and later integrated into the hierarchy. Each sub-policy itself might loss some interpretability if the sub-problem is difficult enough to require a deep learning. By contrast, each of our neurons will maintain its meaning regardless of the complexity, although our neural network could become too complex as well for difficult problems.

\textbf{From interpreable DRL to SRD}. Our model is called \textit{self reward} design because our robot computes its own reward, similar to DRL computation of Q-values. However, human design is necessary to put constraints on how self-rewarding is performed so that interpretability is maintained. In our SRD framework, human designer has the responsibility of understanding the problems, dividing the problems into smaller chunks and then finding the relevant modules to plug into the design in a fully interpretable way (see our first example in which we use convolution layer to create the \textit{food location detector}). We intend to take interpretable DRL to the extreme by advocating the use of very fine-grained, semantically meaningful components.

Other relevant concepts include for example self-supervised Learning. DRL like Value Prediction Network (VPN\cite{NIPS2017_ffbd6cbb}) is self-supervised. Exploration-based RL algorithm is applied i.e. the model gathers data real-time for training and optimization on the go. Our model is similar in this aspect. Unlike VPN, however, our design avoids all the abstraction of DNN i.e. ours is interpretable. Our SRD training is also self-supervised in the sense that we do not require datasets with ground-truth labels. Instead, we induce semantic bias via interpretable components to achieve the correct solutions. The components possess trainable components, just like DRL, and we demonstrate that our models are thus also able to achieve high performance. Our pipeline includes several rollouts of possible future trajectories similar to existing RL papers that use imagination components, with differences as the following. Compared to uncertainty-driven optimization\cite{pmlr-v78-kalweit17a} towards a target value \(\hat{y}_i=r_i+\gamma Q'_{i+1}\) (heavily abbreviated), SRD (1) is similar because agent updates on every imaginary sample available, but (2) has different, context-dependent loss computation. 

\section*{Novelty and Contributions}
We introduce a hybrid solution to reinforcement learning problem: SRD is a deliberately interpretable arrangement of trainable NN components. Each component leverages the tunability of parameters that has helped DNN achieve remarkable performance. The major novelty of our framework is its full interpretability through component-wise arrangement of our models. Our aim is to encourage the development of RL system with both high interpretability and optimizability via the use of tunable parameters of NN components.

\textbf{General framework with interpretable components for specialized contexts}. In general, SRD uses a neural network for the robot/agent to choose its actions, plus the pre-frontal cortex as the mechanism for self reward; this is somewhat comparable to the actor-critic setup. There is no further strict cookie-cutter requirement for SRD since fine-grained interpretability (as we defined previously) may need different components. However, generally, the components only differ in terms of arrangement i.e. existing, well-known components such as convolution layers are arranged in meaningful ways. In this paper, we present the following interpretable components:
\begin{enumerate}[leftmargin=*,topsep=0pt]
\itemsep0em
\item stimulus-specific neuron. With a combination of manually selected weights for convolutional kernel or fully connected layers plus the activation functions (e.g. selective or threshold), each neuron is designed to respond to very specific situation, thus greatly improving interpretability e.g. Food Location Detector in robot fish, eq. \ref{eq:act}. Named neurons such as \(fh, ft\) in robot fish example and \(PG, SZ\) in fish sale auction example ensure that the role and meaning of these neurons are fully understandable.
\item The self-reward design (SRD). In a mammalian brain, prefrontal cortex (PFC) manages internal decision\cite{Miller2002}. Our models are equipped with interpretable PFC modules designed to provide interpretability e.g. we explicitly see why robot fish decides to move rather than eat. We demonstrate how different loss functions can be tailored to the different scenarios.
\item Other components for such as ABA and DeconvSeq for lavaland problem can be found in the appendix.
\end{enumerate}

\textbf{So, what's the benefit of our model? Interpretability and efficiency}. We build our interpretable designs based on existing DNN modules i.e. we leverage the tunable parameters that make Deep RL works. The aim is to achieve both interpretability and good performance without arbitrary specification of reward (like pre-DRL models). \textit{Furthermore, targeted design is efficient}. Our SRD models only use modules from standard DNN modules such as convolution (conv), deconvolution (deconv) and fully-connected (FC) layers with few trainable parameters (e.g. only 180 parameters in Robot2NN). With proper choice of initial parameters, we can skip the long, arduous training and optimization processes that are usually required by DRL models to learn unseen concepts. We trade off the time spent on training algorithm with the time spent on human design, thus addressing what is known as the \textit{sample inefficiency}\cite{8460655} (the need for large dataset hence long training time) in a human-centric way.

\begin{figure}[h]
\begin{center}
\includegraphics[width=0.7\textwidth]{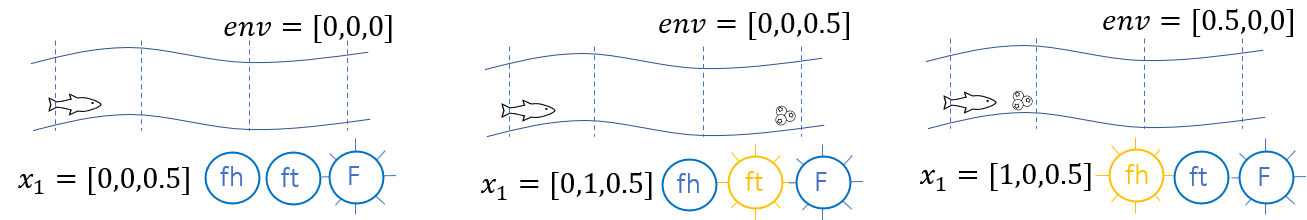}
\end{center}
\caption{Robot Fish setting. The fish is half full/hungry, \(F=0.5\). Left: no food. Middle: food on \(env_3\), thus ``food there" neuron lights up. Right: food on \(env_1\), thus ``food here" neuron lights up.}
\label{fig:fish}
\end{figure}

\section*{Robot fish: 1D toy example}
\label{section:robotfish}
\textbf{Problem setting}. To broadly illustrate the idea, we start with a one-dimensional model Fish1D with Fish Neural Network (FishNN) deliberately designed to survive the simple environment. Robot Fish1D has \textit{energy} which is represented by a neuron labelled \(F\). Energy diminishes over time. If the energy reaches 0, the fish dies. The environment is \(env=[e_1,e_2,e_3]\) where \(e_i=0.5\) indicates there is a food at position \(i\) and no food if \(e_i=0\). The fish is always located in the first block of \(env\), fig. \ref{fig:fish}(A). In this problem, `food here' scenario is \(env=[0.5,0,0]\) which means the food is near the fish. Similarly, `food there' scenario is \(env=[0,0.5,0]\) or \(env=[0,0,0.5]\), which means the food is somewhere ahead and visible. `No food' scenario is \(env=[0,0,0]\).

\textbf{Fish1D's Actions}. (1) `eat': recover energy \(F\) when there is food in its current position. (2) `move': movement to the right. In our implementation, `move' causes \(env\) to be rolled left. If we treat the environment as an infinite roll tape and \(env\) as fish vision's on the 3 immediately visible blocks, then the food is available every 5 block.

\textbf{How to design an interpretable component of neural network?} First, we want the fish to be able to distinguish 3 scenarios previously defined: food here, food there and no food. Suppose we want a neuron that strongly activates when there is food nearby (name it \textit{food here} neuron, \textit{fh}), another neuron that strongly activates when there is a food nearby (name it \textit{food there} neuron, \textit{ft}), and we want to represent the no-food scenario as `neither \textit{fh} and \textit{ft} respond'. How do we design a layer with two neurons with the above properties? We use 1D convolution layer and \textit{selective activation} function as \(\sigma_{sa}(x)=\epsilon/(||x||^2+\epsilon)\), as the following.

\textbf{The \textit{fh} and \textit{ft} neurons}. Define the activation of \textit{fh} neuron as \(a_{fh}= \sigma_{sa}[conv_{fh}(env)]
\) where \(conv_{fh}\), a Conv1D with weight array \(w_{fh}=[1,0,0]\) and bias \(b_{fh}=0.5\). When there is food near the fish, we get \(y_{fh}=conv_{fh}(env)=[1,0,0]* [0.5,0,0]-0.5=0\) where \(*\) denotes the convolution operator, so \(a_{fh}=\sigma_{sa}(y_{fh})=1\). This is a strong activation of neuron, because, by design, the maximum value of selective activation function is \(1\). We are not done yet. Similar to \(a_{fh}\) above, define \(a_{ft}\). The important task is to make sure that when `there is food there but NOT HERE', \(a_{ft}\) activates strongly but \(a_{fh}\) does not. They are \(w_{ft}=[0,1,1],b_{ft}=-0.5\). Together, they form the first layer called the Food Location Detector (FLD). Generally, we have used
\begin{equation}
\label{eq:act}
a_\eta = \sigma_{sa}[conv_{\eta}(env)]
\end{equation}

\begin{figure}[h]
\begin{center}
\includegraphics[width=0.9\textwidth]{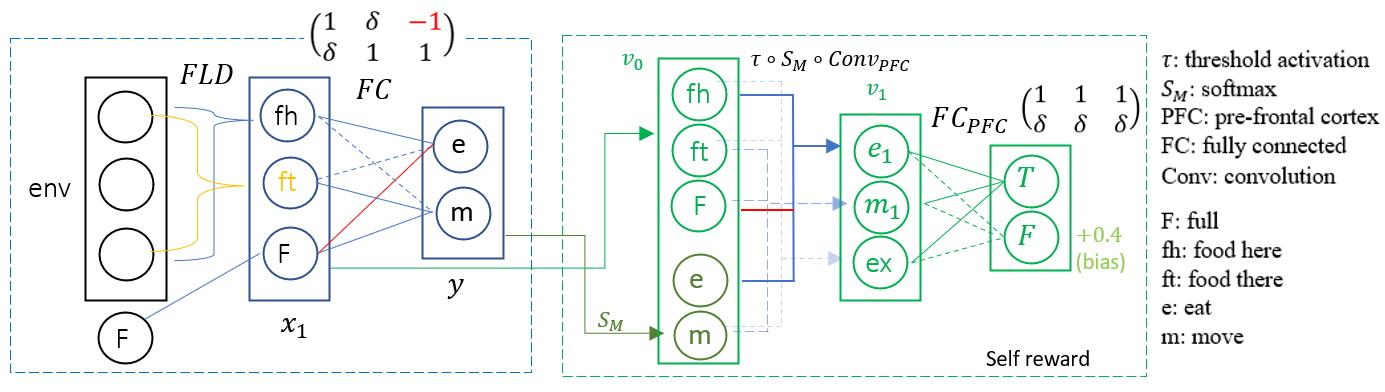}
\end{center}
\caption{FishNN architecture, \(\delta>0\) a small value.}
\label{fig:fishnn}
\end{figure}

\textbf{Interpretable FishNN}. To construct the neural network responsible for the fish's actions (eat or move), we need one last step: connecting the neurons plus fish's internal state (energy) (altogether \([a_{fh},a_{ft},F]\)) to the action output vector \([eat,move]\equiv [e,m]\) through FC layer, as shown in fig. \ref{fig:fishnn} blue dotted box. The FC weights are chosen meaningfully e.g. `eat when hungry  and there is food' and to avoid  scenarios like `eat when there is no food'. This is interpretable through manual weight and bias setting.

\begin{figure}[h]
\begin{center}
\includegraphics[width=0.7\textwidth]{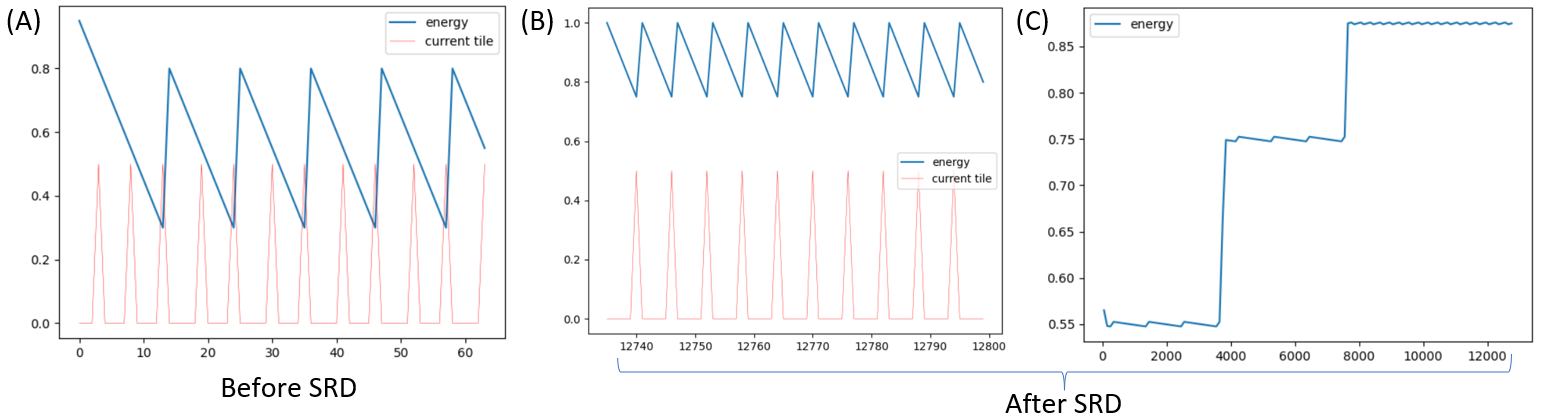}
\end{center}
\caption{(A) plot of energy (\(F\), blue) and food availability (red) for untrained model. SRD-trained model at its early stage looks almost identical. (B) same as (A) but after 12000 SRD training iterations. (C) Average energy of SRD-trained fish model.}
\label{fig:fishresult}
\end{figure}

\textbf{Is FishNN's decision correct?} The prefrontal cortex (PFC) decides whether FishNN's decision is correct or not. PFC is seen in fig. \ref{fig:fishnn} green dotted box. The name `PFC' is only borrowed from the neuroscience to reflect our idea that this part of FishNN is associated with internal goals and decisions, similar to real brain\cite{Miller2002}. How do we construct PFC? First, define \textit{threshold activation} as \(\tau(x)=Tanh(LeakyReLU(x))\). Then PFC is constructed deliberately in the same way FishNN is constructed in an interpretable way as the following. 

\begin{table}[th]
\caption{Weights and biases of Robot Fish's convolution layer}
\begin{center}
\begin{tabular}{ c c c }
  Output & Conv weights & bias \\
\hline
 \(a_{fh}\)& \([1,0,0]\)& \(-0.5\)\\
 \(a_{ft}\)& \([0,1,1]\)& \(-0.5\)\\
 \(e_1\) & \([1,0,-1,1,0]\) & None \\ 
 \(m_1\) & \([0,1,-1,0,1]\) & None\\  
 \(ex\) & [-1,-1,0,0,1] & None \\    
\end{tabular}
\label{table:weightfish}
\end{center}
\end{table}

First, aggregate states from FishNN into a vector \(v_0=[a_{fh},a_{ft},F,e,m]\). This will be the input to PFC. Then, \(v_0\) is processed using \(conv_{PFC}\) followed by softmax and threshold activation. The choice of weights and biases can be seen in table \ref{table:weightfish}. With this design, we achieve meaningful activations \(v_1=[e_1,m_1,ex]\) as before. For example, \(e_1\) is activated when ``there is food and the fish eats it when it is hungry", i.e. \(fh=1,F<1\) and \(e\) is activated relative to \(m_1\). The output of PFC is binary vector \([True,False]=[T,F]\) obtained from passing \(v_1\) through a FC layer \(FC_{PFC}\). In this implementation, we have designed the model such that the activation of any \(v_1\) neuron is considered a \textit{True} response; otherwise it is considered false. This is how PFC judges whether FishNN's action decision is correct. 

\textbf{Self reward optimization}. As seen above, fish robot has FishNN that decides on an action to take and the PFC that determines whether the action is correct. Is this system already optimal? Yes, if we are only concerned with the fish's survival, since the fish will not die from hunger. However, it is not optimal with respect to average energy. We optimize the system through standard DNN backpropagation with the following self reward loss
\begin{equation}
\label{eq:srdloss}
loss=CEL(z,argmax(z))
\end{equation}
where CEL is the Cross Entropy Loss and \(z=\Sigma_{i=1}^{mem} [T,F]_i\) is the accumulated decision over \(mem=8\) iterations to consider past actions (pytorch notation is used). The ground-truth used in the loss is \(argmax(z)\), computed by the fish itself: hence, self-reward design.

\textbf{Results}. Human design ensures survival i.e. problem is solved correctly. Initially, fish robot decides to move rather than eat food when \(F\approx 0.5\), but after SRD training, it will prefer to eat whenever food is available, as shown in fig. \ref{fig:fishresult}. New equilibrium is attained: it does not affect the robot's survivability, but fish robot will now survive with higher average energy.

\section*{Fish Sale Auction}
\label{section:fishsale}
Here, we introduce a more complex scenario that justifies the use of SRD: the fish sale auction. In this scenario, multiple agents compete with each other in their bid to purchase a limited number of fish. A central server manages the dynamic bid by iteratively collecting agents' decision to purchase, to hold (and demand lower price) or to quit the auction. Generally, if the demand is high, the price is automatically marked up and vice versa until all items are sold out, every agent has successfully made a purchase or decided to quit the auction, or the termination condition is reached. In addition, the agents are allowed to fine-tune themselves during part of the bidding process, for example to bias themselves towards purchase when the price is low or when the demand is high.

\textbf{Interpretability requirement}. Here is probably the most important part. The participating agents are required to submit automated models that make the decision to purchase a fish or not based on its price and other parameters (e.g. its length and weight). The automated models have to be interpretable, ideally to prevent agents from submitting malicious models that sabotage the system e.g. by artificially lowering the demand so that the price goes down. Interpretability is required because we want a human agent to act as a mediator, inspecting all the models submitted by the participants, rejecting potentially harmful models from entering the auction. 

Deep RL models and other black box models are not desirable in our auction model since they are less amenable to meaningful evaluation. As we already know, deep RL models have become so powerful they might be fine-tuned to exploit the dynamic of a system, and this will be difficult to detect due to their black-box nature. Furthermore, unlike Mujoco and Atari games, there is no reward to maximize in this scenario, especially because the dynamic depends on many other agents' decision. In other words, standard deep RL training may not be compatible with this auction model since there is no true ``correct" answer or maximum reward to speak of. Our SRD framework, on the other hand, advocates the design of meaningful self-reward; in our previous 1D robot fish example, PFC is designed to be the interpretable cognitive faculty for the agent to decide the correctness of its own action.

\textit{Remark}. This scenario is completely arbitrary and is designed for concept illustration. The item being auctioned can be anything else, and, more importantly, the scenario does not necessarily have to be an auction. It could be a resource allocation problem in which each agent attempts to justify their need for more or less resources, or it could be an advertising problem in which each agent competes with each other to win greater exposure or air time.

\subsection*{The Server and Fish Sale Negotiator}
Now we describe the auction system and a specific implementation. First, we assume that all agents follow the SRD framework and no malicious models are present, i.e. any undesirable models have been rejected after some screening process. In practice, \textit{screening} is a process whereby human inspector(s) is/are tasked to read the agents' model specifications and manually admit only semantically sensible models. This inspection task is not modeled here; instead, a dummy screener is used to initiated interpretable SRD models by default -we will simulate a system with failed screening process later for comparison.
\begin{figure}[h]
\begin{center}
\includegraphics[width=\textwidth]{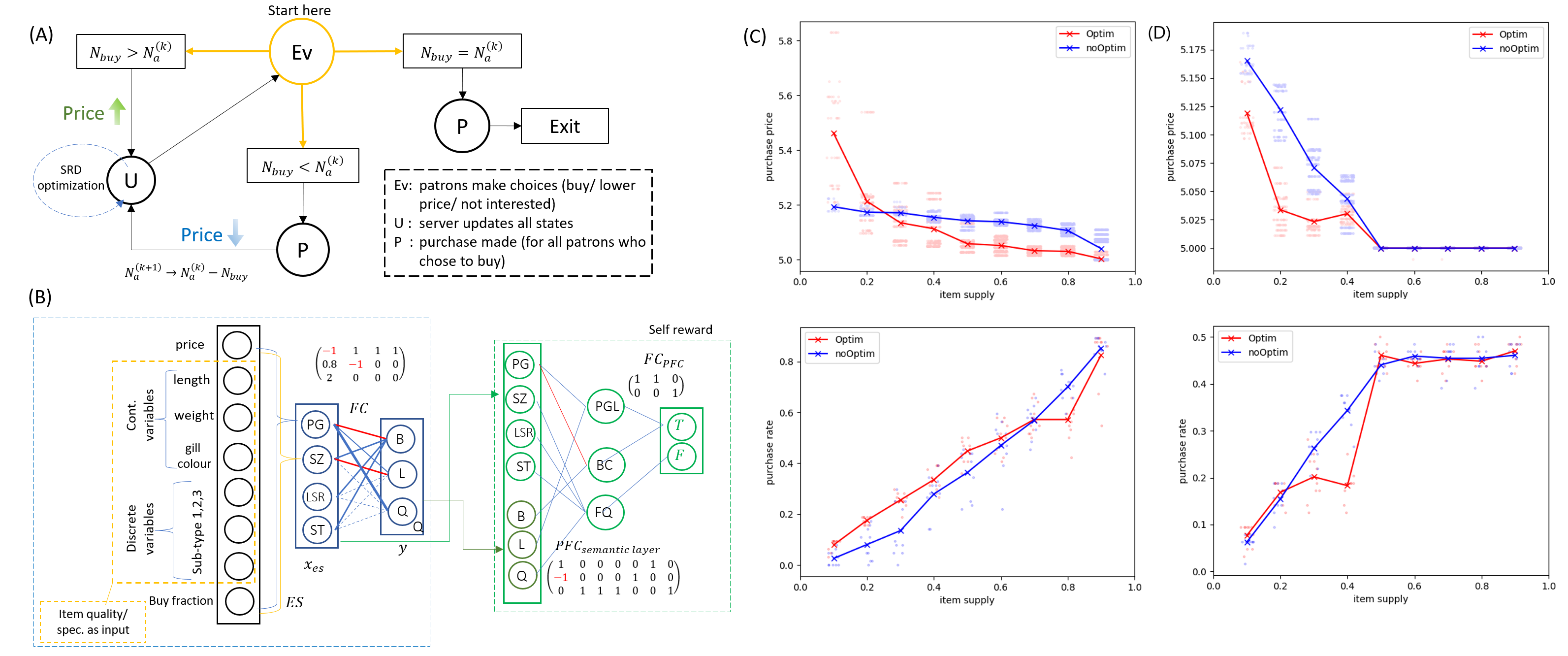}
\end{center}
\caption{(A) Fish sale server. Ev: evaluation process. U: the states update process. P: updating records of successful purchases. (B) The Fish Sale Negotiator. (C) Optim (noOptim) denotes fish sale auction proceeding in which (no) SRD optimization is performed. Each dot in the background corresponds to an actual purchase price at a given \textit{item supply}, whose value in the plot is slightly perturbed to show multiple purchases at similar prices. \textit{Top}: purchase price vs item supply. With SRD optimization, the inverse trend (red) is more pronounced i.e. when there are much fewer fish available relative to the no. of patrons, the fish tends to be sold at a higher price. \textit{Bottom}: purchase rate vs item supply, where purchase rate is the fraction of available fish sold. (D) Same as (C) but half the participants submit malicious models to the auction.}
\label{fig:fishauction}
\end{figure}

The server's state diagram is shown in fig. \ref{fig:fishauction}(A). Data initialization (not shown in the diagram) is as the following. The main item on sale, the fish, is encoded as a vector \((p,l,w,g,st_1,st_2,st_3,f)\in X\subseteq \mathbb{R}^8\) where \(p=5\) denotes the price, \(l,w,g\) are continuous variables, respectively the length, weight, and gill colour normalized to \(1\) (higher values better), \(st_i\) discrete variables corresponding to some unspecified sub-types that we simply encode as either \(-0.5\) or \(0.5\) (technically binary variables) and \(f\) the fraction of participants who voted to purchase in the previous iteration (assumed to be \(0.5\) at the start). The base parameter is encoded as \((5,1,1,1,0.5,0.5,0.5,0.5)\), and for each run, 16 variations are generated as the perturbed version of this vector (data augmentation). 

The server then collects each agent's decision to purchase, hold or quit (corresponding to Ev in the diagram or \textit{\_patrons\_evaluate} function in the code) and redirects the process to one of the three following branches. If the demand at the specific price and parameters is high, i.e. \(N_{buy}>N_a^{(k)}\) where \(N_a^{(k)}\) denotes the remaining number of fish available at the k-th iteration, then the states are updated, including price increment. If the demand is low, then purchase transaction is made for those who voted to buy, availability reduced \(N_a^{(k+1)}=N_a^{(k)}-N_{buy}\), price lowered, and states are correspondingly updated. If the number of purchase is equal to the number of available fish, the purchase is made and the process is terminated. The above constitutes one iteration in the bidding process and this process is repeated up to a maximum of 64 times. In our experiments we observe that the process terminates before this limit is reached, often with a few fish unsold.
 
The Fish Sale Negotiator (FSN) is shown in fig. \ref{fig:fishauction}(B) and this is the fully interpretable SRD model that makes the purchase decision. Each agent will initialize one SRD model with fully interpretable weights, i.e. each agent adjusts their model's parameters based on their willingness to purchase. In our implementation, we add random perturbation to the parameters to simulate the variation of participants. The full details are available in the github, but here we describe how this particular model achieves fine-grained interpretability. FSN is a neural network with (1) the external sensor (ES) module that takes in the fish price and parameters as the input and compute \(x_{es}\) and (2) a fully-connected (FC) layer that computes the decision \(y\). 

\textit{External sensor (ES) module} is a 1D convolution layer followed by threshold or selective activations. This module takes the input \(x\in X\) and outputs \(x_{es}\in\mathbb{R}^4\). The weights \(w_{ES}\) and biases \(b_{ES}\) of convolution layer are chosen meaningfully as the following (note that its shape is \((4,1,8)\) based on pytorch notation). For example, \(w_{ES}[0,:,8]\) corresponds to the PG neuron, which is a neuron that lights up more strongly when the price is higher than baseline and, to a lesser extent, when the continuous variables length, weight and gill colour are lower than \(1\). Thus, \(w_{ES}[0,:,8]=(1/b,-2\delta,-2\delta,-2\delta, 0,0,0,0)+\delta\) where \(b=5\) is the baseline purchase price and \(\delta=0.001\) is a relatively small number. The bias corresponding to \(PG\) is set to \(0\). 

Likewise, other neurons SZ, LSR and ST are initialized similarly: (1) SZ scales with fish sizes i.e. length and weight, thus \(w_{ES}[1,:,8]=(0,1/2,1/2,0,0,0,0,0)+\delta\) with zero bias (2) LSR with the fraction of purchase in the previous iteration (LSR stands for limited supply rush); semantically, this means that the model is more likely to make the decision to purchase if the demand is high (so that it is not outbidden by competitors) thus \(w_{ES}[2,:,8]=(0,0,0,0,0,0,0,1)+\delta\) with zero bias (3) \(ST\) focuses on the specific sub-type \((0.5,0.5,0.5)\), i.e. the neuron is less activated when the sub-type does not match, e.g. \((0.5,-0.5,0.5)\) thus \(w_{ES}[3,:,8]=(0,0,0,0,1/3,1/3,1/3,0)+\delta\) with \(-0.5\) bias. To see how these numbers are adjusted, refer to how we run our codes with \textit{-{}-stage observation} argument.

Implicit augmentation is performed by cloning \(D_{ic}\) copies of the input \(x\in X\) and stacking them  
to \((p',l',...f')\in \mathbb{R}^{8\times D_{ic}}\) where \(p'=(p,p_1,...,p_{D_{ic}})\) (likewise the other variables) and \(D_{ic}=5\) is called the \textit{implicit contrastive dimension} (the number is arbitrarily chosen in this experiment). This is performed as a form of data augmentation in similar spirit to contrastive learning\cite{chen2021exploring, chen2020simple, chen2020improved}, in which the cloned copies are randomly perturbed. The \(ES\) module will have \(dilation=D_{ic}\) so that it outputs \(D_{ic}\) instances of each of the aforementioned neurons that later can be averaged. The benefit of implicit augmentation is clearer in practical application. Suppose the server resides in a remote location and each agent sends and receives its model repeatedly during fine-tuning. If data augmentation is performed in the server and sent to the agent, the network traffic load will be greater. Instead, with implicit augmentation, the process is more efficient since there will be less data transferred back and forth.

\textit{Fully-connected layer}, FC, is designed similarly. Taking \(x_{es}\) at the input, this layer computes the activation of neurons \(B,L,Q\), respectively buy, hold (and lower price) and quit. The decision is made greedily using torch.argmax. We can see that the layer is semantically clear: neuron PG contributes negatively to buy decision, since buying at higher price is less desirable. Neuron SZ contributes positively to neuron B since longer or heavier fish is more desirable; others can be explained similarly. In our experiments, the bias values of FC layer are initiated with some uniform random perturbation to simulate a variation in the strength of intent to purchase the fish.

\textit{Prefrontal Cortex} (PFC) and SRD optimization. PFC performs the self-reward mechanism, just like 1D robot fish PFC. The neurons are semantically meaningful in the same fashion. PGL neuron strongly activates the true \(T\) neuron when PG and L are strongly activated; this means that deciding to hold (L activated) when the price is high (PG activated) is considered a correct decision by this FSN model. BC neuron corresponds to buying at low price, a desirable decision (hence this activates \(T\) as well) while FQ neuron corresponds to ``false quitting", i.e. the decision to quit the auction when SZ, LSR and ST are activated (i.e. when the fish parameters are desirable) is considered a wrong decision, hence activating \(F\) neuron. With this, SRD optimization can be performed by minimizing loss like equation \ref{eq:srdloss}. In this particular experiment, optimization setting is randomized from one agent to another and hyperparameters are arbitrarily chosen. The no. of epochs are chosen uniformly between 0 to 2 (note that this means about one third of the participants do not wish to perform SRD optimization), batch size is randomly chosen between 4 to 15 and learning rate on a standard Stochastic Gradient Descent is initiated uniform randomly \(lr\in [10^{-7},10^{-4}]\). Also, SRD optimization is allowed only during the first 4 iterations (chosen arbitrarily).

\subsection*{Auction results}
Each auction in this paper admits \(n=64\) participants given an \textit{item supply} or rarity \(r\). In our code, \(r\) is defined such that the no. of available fish on auction corresponds to \(r\times n\). Each such auction is repeated 10 times and each trial independent of any other. The results of each trial is then collected, and the mean values of purchase price and purchase rate are shown in fig. \ref{fig:fishauction}(C). In this specific implementation, the auction that allows SRD optimization shows a more pronounced price vs supply curve, as shown in fig. \ref{fig:fishauction}(C) top, red curve. We can see that lower supply results in higher purchase price on average. Fig. \ref{fig:fishauction}(C) bottom shows that the fraction of successful buys varies in a nearly linear fashion as a function of supply.

We should remember that the well-behaved trends that we just observed are obtained from \textit{sensible} automated models, in the sense that they make decisions with humanly understandable reasoning such as ``buy when the price is low". This is possible thanks to the full interpretability afforded by our SRD framework. For comparison, we are interested in the case where interpretability is lacking and malicious actors sneak into auction. Thus we repeat the experiments, except half the agents always make \textit{hold} decisions, intending to sabotage the system by forcefully lowering the price. The results are shown in fig. \ref{fig:fishauction}(D): the graph of purchase price (top) is greatly compromised which translates into a loss for the auction host. The graph of purchase rate (bottom) appears to be slightly irregular at low item supply. However, the purchase rate plateaus off at higher supply (malicious agents refuse to make any purchase) i.e. the auction fails to sell as many fish as it should have, resulting in a loss. By admitting only sensible interpretable models, the auction host can avoid such losses.

\section*{More scenarios}
Generally, SRD framework encourages developers to solve control problems with neural networks in the most transparent manner. Different problems may require different solutions and there are possibly infinitely many solutions. This paper is intended to be a non-exhaustive demonstration of how common components of black-box neural network can be repurposed into a system that is not black-box. Here, we will briefly describe two more scenarios and leave the details in the appendix: (1) 2D robot lavaland and (2) the Multi-Joint dynamics with Contact (MuJoCo) simulator.

\subsection*{2D robot lavaland}
\begin{figure}[h]
\begin{center}
\includegraphics[width=1\textwidth]{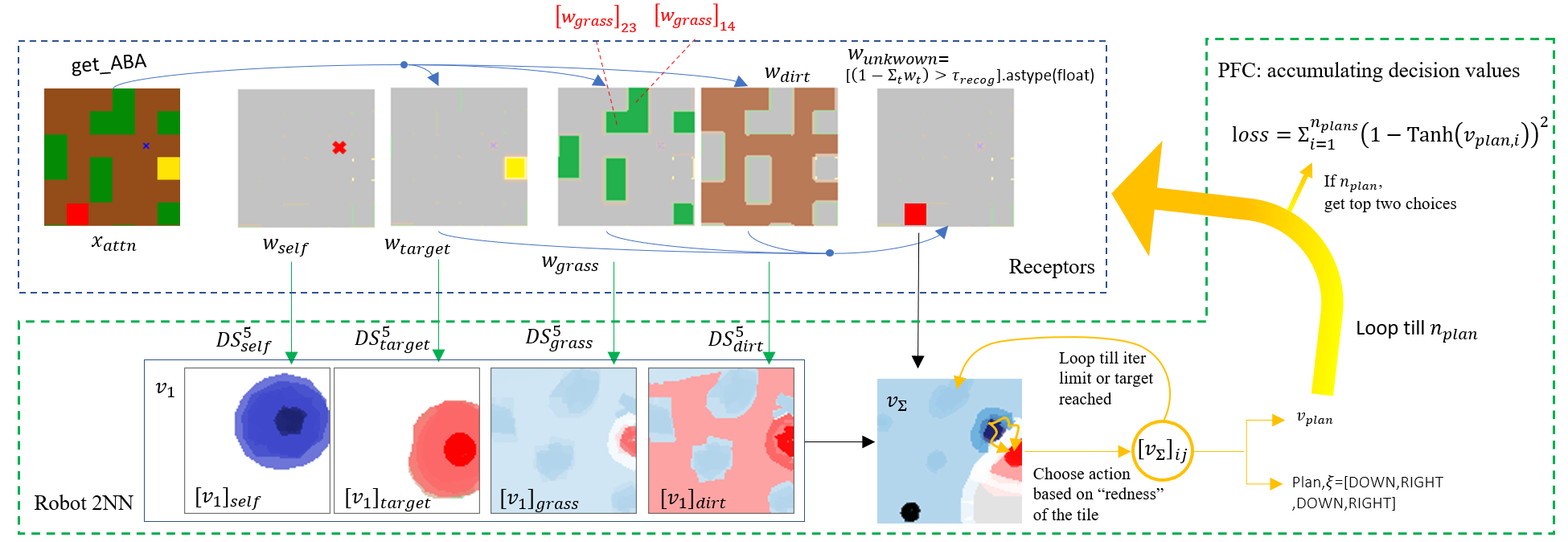}
\end{center}
\caption{Robot2NN schematic. Receptors help split incoming signals for further processing. \(x_{attn}\) (the whole visible map) and \(w_{self}\) (the agent's position) are the direct input to the model. In \(v_1,v_\Sigma\), red values are positive (desirable), white zero and blue negative (not desirable). PFC (green dotted box) contains trainable series of parameters to make and adjust decisions. \([w_{grass}]_{14},[w_{grass}]_{23}\) are examples of strong activations that are interpretable through tile-based module.}
\label{fig:robot}
\end{figure}

Like Dylan's IRD paper, we use lavaland as a test bed. An agent/robot (marked with blue x) traverses the \textit{lavaland} by moving from one tile to the next towards the objective, which is marked as a yellow tile. With components such as ABA and selective activation function (see appendix), semantically meaningful neurons are activated as we have done in our previous two examples. In this scenario, each neuron will respond to a specific tile in the map. More specifically, brown tiles (dirt patches) are considered easy to traverse while green tiles (grassy patches) are considered harder to traverse. The robot is designed to prefer the \textit{easier} path, thus each neuron responds more favourably towards brown tiles as shown by red patches of \(v_1\) in fig. \ref{fig:robot}. The problem is considered solved if the target is reached within 36 steps

Furthermore, we demonstrate how \textit{unknown avoidance} can be achieved. There will be red tiles (lava) that are not only dangerous, but also have never been seen by the robot during the training process (thus unknown). The agent is trained only on green and brown tiles, but when it encounters the unknown red tiles, our interpretable design ensures that the agent avoids the tile, thus \textit{unknown avoidance}. This is useful when we need the agent to ``err on the safer side" basis. Once human designer understands more about the unknown, using SRD design principle, a new model can be created by taking into account this unknown. The full experiments and results are available in the appendix.

\subsection*{MuJoCo with SRD}
MuJoCo\cite{mujoco6386109} is a well-known open source physics engine for accurate simulation. It has been widely used to demonstrate the ability of RL models in solving control problems. Here, we briefly describe the use of SRD framework to control the motion of a half-cheetah. All technical details are available in the appendix and our github. More importantly, in the appendix, we will describe the design process step by step from the start until we arrive at the full design presented here.

To solve this problem in its simplest setting, an RL model is trained with the goal of making the agent (half-cheetah) learn how to run forward. Multiple degrees of freedom and coordinates of the agent's body parts are available as input and feedback in this control scenario, although only a subset of all possible states will enable the agent to perform the task correctly. More specifically, at each time step, the agent controls its actuators (6 joints of the half cheetah) based on its current pose (we use x and z position coordinates relative to the agent's torso), resulting in a small change in the agent's own pose. Over an interval of time, accumulated changes result in the overall motion of the agent. The objective is for the agent to move its body parts in a way that produces regular motions i.e it runs forward without stalling, tripping, collapsing etc. Typically, this is achieved by a training (maximizing some rewards); without proper training, the agent might end up stuck in an unnatural pose or even start moving backwards.

\begin{figure}[h]
\begin{center}
\includegraphics[width=1\textwidth]{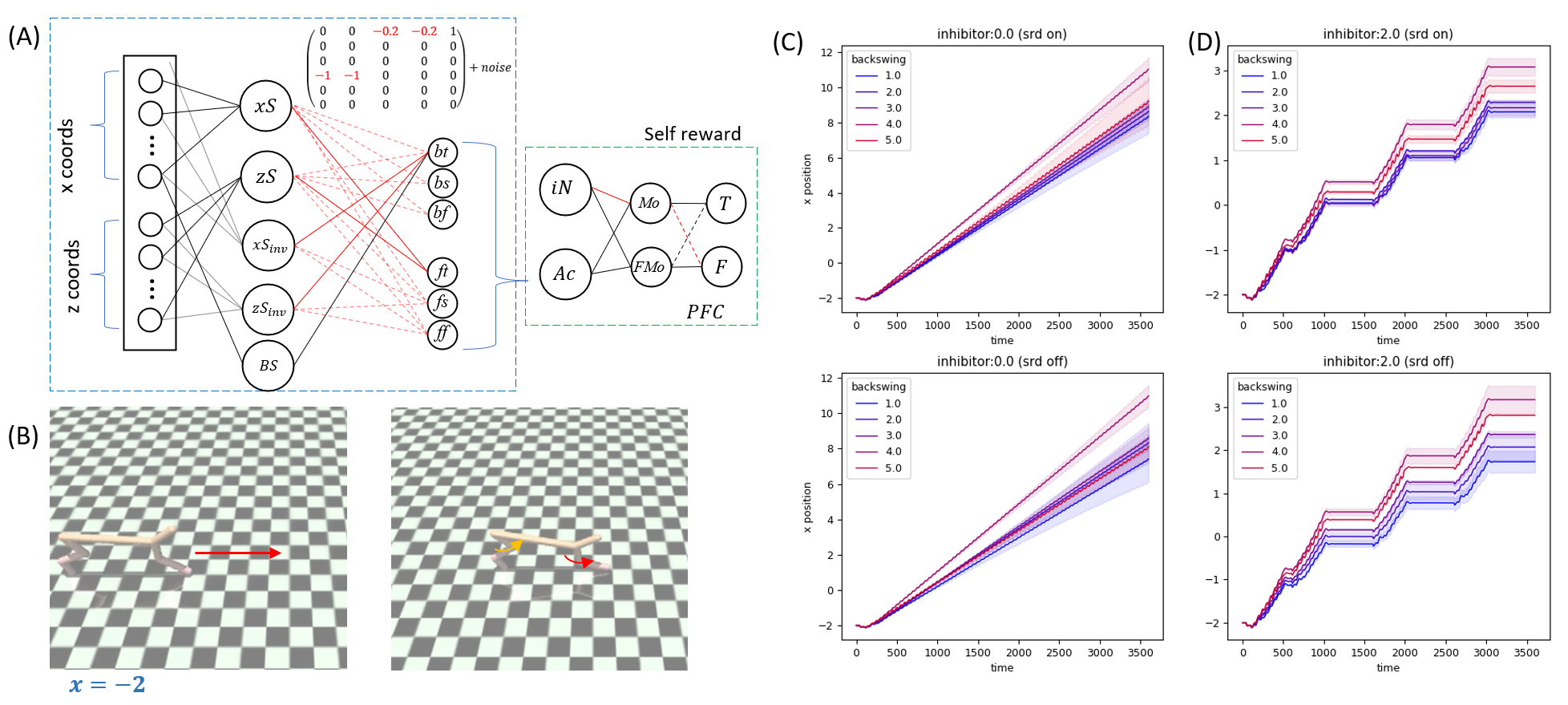}
\end{center}
\caption{(A) The HalfCheetahSRD, the neural network model used to move MuJoCo half-cheetah. (B) Half cheetah starts at position \(x=-2\). Red/orange curved arrows denote the swings of front/rear thighs. (C) mean x position over time for different magnitudes of \textit{backswing}. (D) Same as (C), but with \textit{inhibitor} set to 2, allowing the agent to response the instruction to stop moving.}
\label{fig:mujoco}
\end{figure}

With SRD, deliberate design of a neural network as shown in fig. \ref{fig:mujoco}(A) enables the cheetah to start running forward without training (or any optimization of reward), just like our previous examples. Self reward mechanism is also similar: PFC part of the neural network will decide the correctness of the chosen action, and optimization can be performed by minimizing cross-entropy loss as well. Snapshots of the half cheetah are shown in fig. \ref{fig:mujoco}(B). The plots of mean x displacement from the original position over time are shown in fig. \ref{fig:mujoco}(C) (top: without optimization, bottom with SRD optimization). The average is computed over different trials initialized with small noises and the shaded region indicates the variance from the mean values. We test different \textit{backswings}, i.e. different magnitudes with which the rear thighs are swung. Some backswings result in slightly faster motion. More importantly, the structure of the neural network yield stable motion over the span of backswing magnitudes that we tested. Furthermore, we also tested an additional feature: half cheetah is designed to respond to the instruction to stop moving (which we denote with \(inhibitor=2\)). The result is shown in fig. \ref{fig:mujoco}(D), in which the instruction to stop moving is given at regular intervals. Finally, the effect of SRD optimization is not immediately clear. At lower backswing magnitudes, the optimization might have yielded a more stable configuration, which is more clearly visible in fig. \ref{fig:mujoco}(D) top (compared to bottom).

\section*{Limitation, Future Directions and Conclusion}
\textit{Generalizability and scalability}. An important limitation to this design is the possible difficulty in creating specific design for very complex problems, for example, computer vision problems with high-dimensional state space. Problems with multitude of unknown variables might be difficult to factor into the system, or, if they are factored in, the designers' imperfect understanding of the variables may create a poor model. Future works can be aimed at tackling these problems. Regardless, we should mention that, ideally, a more complex problem can be solved if a definite list of tasks can be enumerated, the robot's state can be mapped to a task and no state requires contradictory actions (or perhaps stochasticity can be introduced). Following the step-by-step approach of our SRD framework, each task can thus be solved with a finite addition of neurons or layers. Further research is necessary to understand how noisy states can be properly mapped to an action.

Also, an existing technical limitation includes the lack of APIs that perform the exact operations we may need to parallelize the imaginative portions of SRD optimization (see lavaland example in the appendix). We have explored implicit contrastive learning for data augmentation but more research is needed to understand the effect of different techniques. Also, a general formula seems to be preferred in the RL field, and this is not currently available in SRD. For now, a standard SRD framework consists of (1) a NN that decides the agent's actions based on its current state and (2) a PFC that facilitates the process of self-reward optimization with true or false output. Further research is necessary.

\textit{Controlling Weights and Biases}. A future study on regularizing weights may be interesting. While we have shown that our design provides a good consistency between the interpretable weights before and after training, it is not surprising that the combination of small differences in weights can yield different final weights that still perform well. So far, it is not clear how different parameters affect the performance of a model, or if there are any ways to regularize the training of weight patterns towards something more interpretable and consistent. 

To summarize, we have demonstrated interpretable neural network design with fine-grained interpretability. Human experts purposefully design models that solve specific problems based on their knowledge of the problems. SRD is also introduced as a way to improve performance on top of the designers' imperfection. With manual adjustment of weights and biases, designers are compelled to assign meaningful labels or names to specific parts, giving the system a great readability. It is also efficient since very few weights are needed compared to traditional DNN.

%\noindent Please note: Abbreviations should be introduced at the first mention in the main text – no abbreviations lists. Suggested structure of main text (not enforced) is provided below.
%
%\section*{Introduction}
%
%The Introduction section, of referenced text\cite{Figueredo:2009dg} expands on the background of the work (some overlap with the Abstract is acceptable). The introduction should not include subheadings.
%
%\section*{Results}
%
%Up to three levels of \textbf{subheading} are permitted. Subheadings should not be numbered.
%
%\subsection*{Subsection}
%
%Example text under a subsection. Bulleted lists may be used where appropriate, e.g.
%
%\begin{itemize}
%\item First item
%\item Second item
%\end{itemize}
%
%\subsubsection*{Third-level section}
% 
%Topical subheadings are allowed.
%
%\section*{Discussion}
%
%The Discussion should be succinct and must not contain subheadings.
%
%\section*{Methods}
%
%Topical subheadings are allowed. Authors must ensure that their Methods section includes adequate experimental and characterization data necessary for others in the field to reproduce their work.

%\bibliography{sample}
%
%\noindent LaTeX formats citations and references automatically using the bibliography records in your .bib file, which you can edit via the project menu. Use the cite command for an inline citation, e.g.  \cite{Hao:gidmaps:2014}.
%
%For data citations of datasets uploaded to e.g. \emph{figshare}, please use the \verb|howpublished| option in the bib entry to specify the platform and the link, as in the \verb|Hao:gidmaps:2014| example in the sample bibliography file.

\section*{Data availability}
Our data and results can all be found in the following: \\\url{https://drive.google.com/drive/u/3/folders/1FoeGgfcO4hdWZynxVFrzPYWvYwIWVZ0p}.

\bibliography{srd_sr}

\section*{Acknowledgements}
This research was supported by Alibaba Group Holding Limited, DAMO Academy, Health-AI division under Alibaba-NTU Talent Program. The program is the collaboration between Alibaba and Nanyang Technological University, Singapore. This work was also supported by the RIE2020 AME Programmatic Fund, Singapore (No. A20G8b0102).

\section*{Author contributions statement}
E.T. performed the experiments. All authors reviewed the manuscript. 

\newpage

{\LARGE Self Reward Design with Fine-grained Interpretability}

{\large Erico Tjoa, Cuntai Guan}

\section*{Appendix}

%To include, in this order: \textbf{Accession codes} (where applicable); \textbf{Competing interests} (mandatory statement). 
%
%The corresponding author is responsible for submitting a \href{http://www.nature.com/srep/policies/index.html#competing}{competing interests statement} on behalf of all authors of the paper. This statement must be included in the submitted article file.
%
%\begin{figure}[ht]
%\centering
%\includegraphics[width=\linewidth]{stream}
%\caption{Legend (350 words max). Example legend text.}
%\label{fig:stream}
%\end{figure}
%
%\begin{table}[ht]
%\centering
%\begin{tabular}{|l|l|l|}
%\hline
%Condition & n & p \\
%\hline
%A & 5 & 0.1 \\
%\hline
%B & 10 & 0.01 \\
%\hline
%\end{tabular}
%\caption{\label{tab:example}Legend (350 words max). Example legend text.}
%\end{table}
%
%Figures and tables can be referenced in LaTeX using the ref command, e.g. Figure \ref{fig:stream} and Table \ref{tab:example}.

\subsection*{This Paper Focuses Heavily on Interpretable Human Design}
\label{appendix:focus}
We reorganize this paper heavily based on ICLR 2022 reviewers' comments and have further revised the majority of this paper based on Scientific Report reviewers' comments. Thanks to them, we understand that our focus on fine-grained interpretability (through direct manipulation of weight and biases) seems to have been lost or easily overlooked in the previous version of our paper. There is a tendency to focus on performance and solution to generalizability. We have answers for this (1) we do achieve high performance, and comparison might be very redundant. Our intention is to compare interpretability instead of performance. (2) Scalability. We naturally start by demonstrating the design on simple problems. Starting with a highly complex problems may be counterproductive and very difficult to present. In this revised paper, we have focused on the clarity of presentation.

\textbf{Convergence}. The design process should ideally require step-by-step testing that ensures the ``convergence" of the model, which we demonstrate through our solution to MuJoCo half-cheetah. The PFC module that facilitates optimization should also be tested to ensure that the conceptual designs are working. Also, stable convergence is no longer guaranteed\cite{lillicrap2015continuous} in most cases when non-linear functions are involved.

Why do the weights and biases remain interpretable after SRD optimization? In our SRD framework, the initial selection of weights and biases are required to work to an extent, which means that the loss w.r.t a given metric of the problem is not very high from the very start, or at least the loss is not as high as randomly initialized model. This perhaps means that the designed weights and biases are already somewhat near a local minimum point. In practice, the SRD optimization probably only finetunes the parameters so that the model moves slightly nearer the local minimum, thus not changing the weights and biases too much.”

\subsection*{Related Works}
\label{appendix:relatedworks}

In Reward Design Problem\cite{Barto2009WhereDR}, it is observed that, given bounded agents, proxy reward function can be more optimal than the true fitness function which is distinct from the proxy. Dylan's IRD, on the other hand, approximates the true fitness function given an observed proxy reward function, assuming the designer is the bounded agent, i.e. the designer is fallible. IRD performs inversion to compute \(P(w=w^*|\tilde{w},\tilde{M})\) from \(P(\tilde{w}|w^*,\tilde{M})\), where \(w\) is the weight of the reward function \(r(\xi;w)\), where \(\xi\) is the trajectory. We cannot fit the description of our model exactly in the same language, since we opt for an interpretable design with non-traditional RL reward. The closest we have to a reward is \(v_{plan}\) in Robot2NN, shown in fig. \ref{fig:robot} indirectly as the values summed through orange arrows in \(v_{\Sigma}\) described in the next paragraph.

\textbf{Comparison with standard Deep RL}. In the main text, we mention that our we use non-traditional reward function. Deep Q-learning loss function \(L_i(\theta_i)\) computes differences between current \(Q\) value (value of the current state-action pair) and its temporal difference (TD)\footnote{see for example \url{https://www.tensorflow.org/agents/tutorials/0_intro_rl}} given by \(r+\gamma max_{a'}Q(s',a',\theta_{i-1})\). The reward \(r\) can be fixed by the designer; this can be seen on OpenAI's official github\footnote{\url{https://github.com/openai/gym/blob/master/gym/envs/classic_control/cartpole.py}}, and the values are then used to update Q, an expected value of total future rewards, technically also a reward. SRD is similar to DQN in this sense: there is a future cumulative reward. In turn, DQN is similar to standard RL except Q is predicted by the neural network. However, SRD does not apply the loss function \(L_i(\theta_i)\) in DQN since we do not specify TD, which is dictated by the reward that explicitly adds into the sum of future reward. As mentioned, the reward value \(r\) in DQN is specified by the designer and thus DQN robot aims to maximize values whose absolute magnitude of worth is known. By contrast, SRD lets the robot measure the worth of every series of actions. Since the target Tanh score is 1, every measurement of worth is relative with respect to the learning process. Even then, when the robot compares the multiple plans it makes at once, it has its own yardstick for rewarding its own decisions. The absolute magnitude of reward based on this yardstick is decided by the robot's own neural network, thus the ``self reward". Finally, our robot self reward works by always undervaluing its perceived reward w.r.t the ideal plan with score 1, as mentioned in the main text.

\textbf{Self-supervision}. Self-supervised DRL is demonstrated on a simulated car that learns from real-time experience\cite{8460655}; the paper also applies the model to a real-world RC car. After hybrid graph-based RL model is incorporated, it travels around avoiding collisions without human intervention. Our robots self-supervise similarly. After the initial interpretable design, 2D Robot moves towards the target, favoring or avoiding different types of tiles without human supervision. In particular, grass tiles are recognizable, but less desirable, while lava tiles are ``not recognized", hence must be completely avoided. Informally, the robots have their own preferences on how to assign the eventual values of the actions with known local consequence, while they are careful about their own ignorance induced by imperfect design (unknown avoidance). 

\textbf{Imagination components}. Unlike an existing rollout\cite{10.5555/3295222.3295320}, each of our SRD rollout consists of a series of asymmetric binary choices \(a_1,a_2\) chosen from \(\{a\in\mathcal{A}\}\) so that \(v(a_1)\ge v(a_2)\), where \(\mathcal{A}\) is the set of actions and \(v\) is any generic function that gives each action a local value. The values will be aggregated into a self-reward. Dreamer\cite{Hafner2020Dream} solves RL problem using only latent imagination where many models (such as reward, action and value models) are specified as probability distributions. By contrast, SRD creates no specific sub-model. All values are just NN activations, and they will be aggregated into impromptu, just-in-time scores, based on which the plans can be greedily chosen.

\subsection*{2D Robot in Lavaland}
\begin{figure}[h]
\begin{center}
\includegraphics[width=0.7\textwidth]{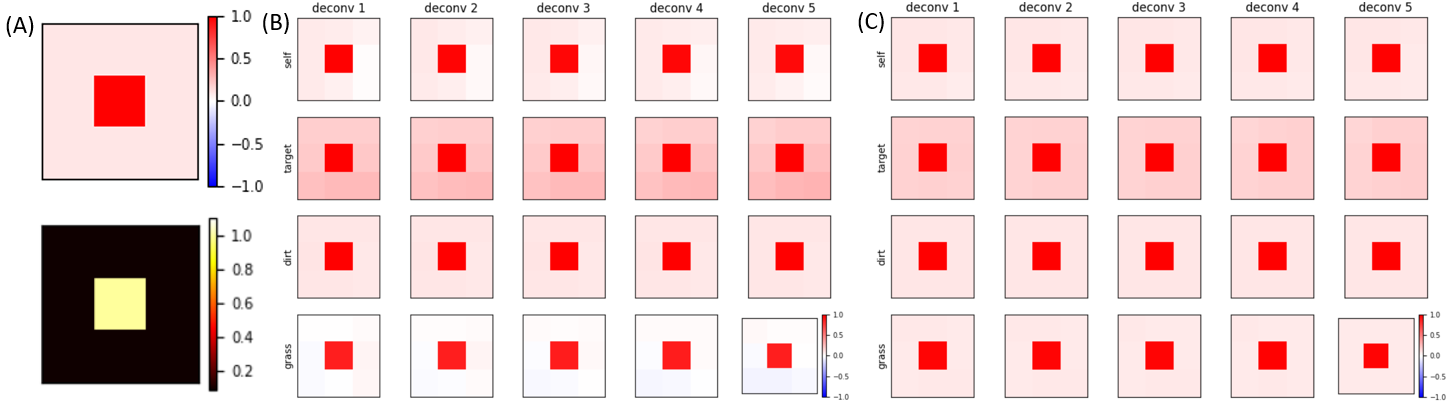}
\end{center}
\caption{(A) Initial parameters; all deconvs in all DeconvSeq are initialized to the same 3x3 weights with center max value of 1 and 0.1 elsewhere. (B) Once trained, variations of weights are observed for Project A expt 1 (all 180 parameters are shown). (C) Similar to (B) but for project Compare A.}
\label{fig:weights}
\end{figure}

In the main text, we briefly discussed 2D robot in the lavaland. We will elaborate further here, starting with the interpretable components used in 2D robot's design:
\begin{enumerate}[leftmargin=*,topsep=0pt]
\itemsep0em

\item The ABA: approximate binary array. With \textit{selective activation} and tile-specific values (colour), we create strong neuron activations that specifically correspond to tile colours. Their visual maps correspond directly to the relevant signal, preserving the ease of readability. 
\item The DeconvSeq. The series of convolutional kernels are intended to provide targeted response in conjunction with ABA e.g. in fig. \ref{fig:robot}, \([v_1]_{target}\) gives strong signal centred around the target. This is done by manually setting the center value of the weights to be higher than the rest (see fig. \ref{fig:weights}). The main selling point is their tunability. The weights are trainable: while each module has been given a specific purpose, e.g. detect target, it is still tunable. We empirically show that the main purpose the kernels' weights are preserved (i.e. center value still highest) after optimization.
\end{enumerate}

\noindent The interpretable design of Robot2NN is shown in the main text fig. \ref{fig:robot}. Tile-based modules in the receptors are designed to respond to different types of stimuli (grass, ground, lava etc); as our previous examples, weights and biases are manually selected. Deconvolutional layers are used in the robot's PFC to give the tiles some scores for the robot to decide its subsequent action (red better, blue worse). More precisely, a stack of deconvolutional layers \(DS^n_t\) (defined later) will be used to create a \textit{favourability gradient}. Robot then chooses an action that generally moves it from blue to red regions. 

Before we proceed, we clarify some of our notational uses. ABA: approximate binary array, an array whose entries are expected to be \(\approx 0\) or \(1\). \(DS^n_{t}\), or DeconvSeq, is the sequence of \(n\) deconvolutional layers for a tile type \(t\). Deconvolutional layer, or deconv, is a regular DNN module. Each deconv is followed by Tanh activation for normalization and non-linearity. Normalization (to magnitude \(1\)) ensures that DeconvSeq compares action choices in relative terms. Tile \(t\) denotes the name of a tile, e.g. grass, but it also denotes its \([0,1]\) normalized RGB value, e.g. for grass, \(t=[0,128,0]/255\) or an array of tile values (they should be obvious from the context). \(\tau_{recog}=10^{-4}\) is the recognition threshold. Designer needs to specify \(P=\{p_{t}:t=target,grass,...\}\). For now \(P\) is the set of untrainable parameters, each \(p_{t}\) roughly acting as the factor for scaling the true reward. They will affect the robot's final preferences for or against different tiles although tunable parameters will accentuate or attenuate them accordingly. They induce the biases that designers input into the model in a simple, interpretable way. We also define the \textit{unknown avoidance} parameter \(u_a\ge 0\).

\textbf{Interpretable tile-based modules} are designed to explicitly map robot's response to each specific tile type. Get\_ABA() function computes the ABA for each tile type: \(w_{t}=\sigma_{sa}\circ\mu[(x_{attn}-t)^2]\) where \(\mu[.]\) is the mean across RGB channel. This is reminiscent of eq. \ref{eq:act}; the difference is, each neuron responds to a tile type at each \(x_{attn}\) coordinates. Neuron activations are computed as \(w_\eta\) where \(\eta=\)target, grass or dirt shown in fig. \ref{fig:robot}, e.g. strong activation for grass detection occurs at \([w_{grass}]_{14}\).

\textbf{Unknown avoidance}. Like Dylan's IRD, we have a reliable mechanism for unknown avoidance, the Boolean array \(w_{unknown}=[(1-\Sigma_{t}w_{t})>\tau_{recog}]\) to be treated as floating point numbers. From the formula, it can be seen that \(w_{unknown}\) aggregates the negation of known activations. The unknown in our case is the lava tile that the imperfect human designer `forgets' to account for.

\textbf{Interpretable tile scores}. The score \(v_\Sigma\) will be used to decide what actions robot will take. It is computed as the following. First, we compute \(v_1\), whose components are \([v_1]_{t}\), as the following. For target tiles and ``self" tiles (original positions), \([v_1]_{t}=p_{t}DS^5_{t}[w_{t}]\). For any other named tiles, we also create the \textit{favourability gradients} based on the initial and target positions, so that, for example, a grass tile nearer the target can be valued more than a dirt tile. Thus, 
\begin{equation}
[v_1]_{t}=p_{t}DS^5_{t}[(w_{target}-w_{self})*w_{t}] 
\end{equation}
Each deconv in DeconvSeq consists of 1 input and 1 output channel. It is fully interpretable through our manual selection of kernel weights: kernel has size 3 with center value \(1\) and side value \(0.1\) as seen in fig. \ref{fig:weights}(A). Fig. \ref{fig:weights}(B, C) show examples of trained weights. This choice of values is intended roughly to create a centred area of effect, where the center of the tile contributes most significantly to \(w_t\); see for example \([v_1]_{target}\) in fig. \ref{fig:robot}. Assign \(v_\Sigma\leftarrow \Sigma_t [v_1]_t\); this value will be dynamically changed throughout each plan-making process.

\textbf{Making one plan}. Each plan is a series of actions, and each action is chosen as the following.  Set the specific values for the target tile and tile of the original position to \(v_0\) and \(-v_0\) respectively, where \(v_0=max\{|v_\Sigma|\}\) is computed without backpropagation gradient to prevent any complications. Finally, to incorporate lava avoidance, or generally avoidance of anything previously unseen, \(v_\Sigma\leftarrow v_\Sigma* (1-w_{unknown})+-u_a * v_0 * w_{unknown}\). From the current position, robot's neighbouring \([v_\Sigma]_i\) where \(i=up, down, left, right\) values are collected, and neighbours with the top two values are chosen. From the top two choices, randomly choose one of them with a 9 to 1 odds, favouring the action with higher value. The randomness is to encourage exploration. After each action, the tile the robot leaves will be assigned \(0.9v_0\) to prevent the robot from oscillating back and forth, where \(v_0\) is separately computed inside step\_update() function. A series of actions are chosen in this manner until the target is reached or a maximum of 36 iterations are reached. Thus we have obtained a \textit{plan} and its score \(v_{plan}=\frac{1}{N_{\xi}}\Sigma_{i,j\in\xi}[v_\Sigma]_{i,j}\), the mean of all values assigned to the chosen tiles where \(\xi\) is the trajectory (see fig. \ref{fig:robot}, orange arrows in \(v_{\Sigma}\)).  

\textbf{Imagining multiple plans for SRD optimization}. Robot makes plans by imagining \(n_{plans}=4\) different trajectories to reach the target. It plans and executes the plan with the highest \(v_{plan}\). Like \cite{pmlr-v78-kalweit17a}, we use all \(n_{plans}\) imagination branches for training (SRD optimization) with possibly novel loss minimization: 
\begin{equation}
\label{eq:one_loss}
loss=\Sigma_{i=1}^{n_{plans}}(1-Tanh(N[v_{plan,i}]))^2
\end{equation}
where \(N[.]\) normalizes \(v_{plan}\) to the magnitude of 1, where normalization factor is computed without gradient to prevent complication.  Thus, robot's PFC always undervalues its reward relative to an abstract, ideal value \(1\). The intended effect is to always try to maximize \(v_{plan}\). A standard Stochastic Gradient Descent with learning rate \(10^{-4}\) is used for optimization. This resembles fish's PFC true or false neurons used for rewarding itself, except it is continuous.

\textbf{Experimental setup and comparison}. We experiment on several initial settings as the following. For each experiment, we randomly generate and save 4096 maps to evaluate the performance of both standard design and SRD trained model, and similarly 4096 maps for SRD training. Note: All codes are included, including codes for creating animated gifs of robot traversing the lavaland.

\textbf{Reaching targets without training, optimized by training}. Table \ref{table:1} shows that even without training (``No SRD" columns), our interpretable designs have enabled relatively high rate of problem solving. With SRD, the accuracies are further improved. 

\begin{table}[h]
\caption{Accuracy comparisons for Project A, project Compare A, project With-Lava-A and project Lava NOAV A. No. 1 to 4 indicate four independent but identical experiments. \(Acc=n_{reached\ target}/4096\). ``No SRD" indicates no training: reasonable accuracy is attainable purely with design. SRD training generally increases the accuracies. Starred values reflect failure modes, matching their respective histograms; see fig. \ref{fig:no_lava} and \ref{fig:lava}.} 
\label{table:1}
\begin{center}
\begin{tabular}{lllllllll}
\multicolumn{1}{c}{ } & \multicolumn{2}{c}{Project A} & \multicolumn{2}{c}{Compare A} & \multicolumn{2}{c}{With Lava A} & \multicolumn{2}{c}{Lava NOAV A}
\\\hline 
\multicolumn{1}{c}{}  &\multicolumn{1}{c}{No SRD} &\multicolumn{1}{c}{ SRD} &\multicolumn{1}{c}{No SRD} &\multicolumn{1}{c}{ SRD} &\multicolumn{1}{c}{No SRD} &\multicolumn{1}{c}{ SRD} &\multicolumn{1}{c}{No SRD} &\multicolumn{1}{c}{ SRD}
\\\hline 
1  & 0.767 & 0.856 & 0.862 & 0.941             & 0.826 & 0.893             & 0.841 & 0.884 \\
2  & 0.765 & 0.845 & 0.871 & 0.589* & 0.821 & 0.893             & 0.850 & 0.909 \\
3  & 0.763 & 0.817 & 0.874 & 0.930             & 0.819 & 0.837             & 0.844 & 0.868\\
4  & 0.760 & 0.868 & 0.863 & 0.422* & 0.823 & 0.762* & 0.835 & 0.907
\end{tabular}
\end{center}
\end{table}

\begin{figure}[h]
\begin{center}
\includegraphics[width=0.8\textwidth]{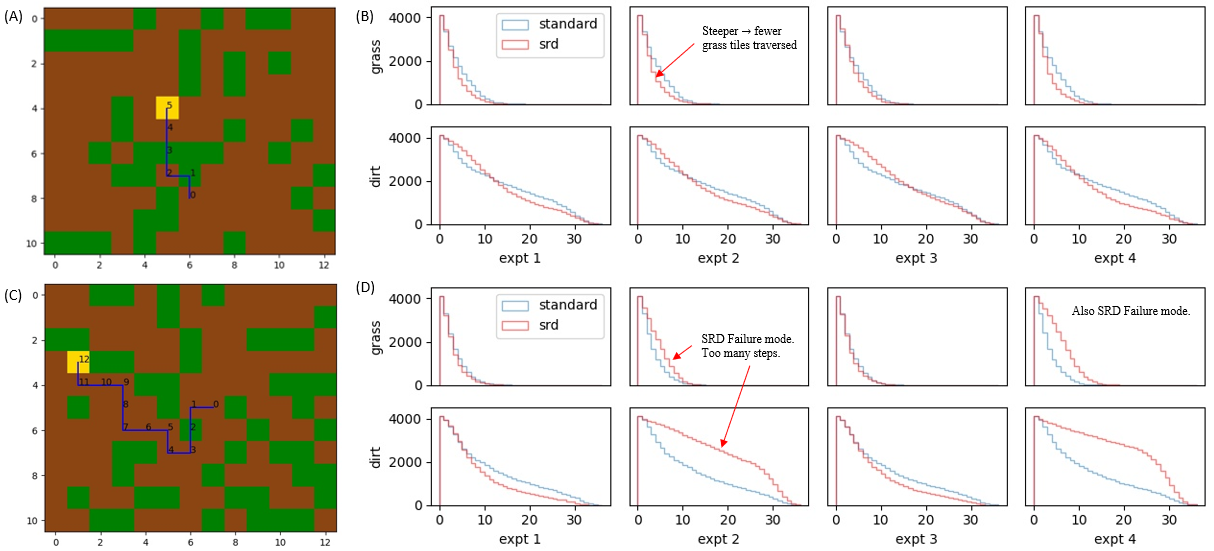}
\end{center}
\caption{(A) A sample trajectory from Project A. (B) Cumulative histogram of the no. of tiles traversed by robot in Project A. Steeper histogram indicates that less tiles of the type are being traversed (for grass, steeper is better) (C) A sample trajectory from Compare A. (D) Similar to B, but for robot in Compare A.}
\label{fig:no_lava}
\end{figure}

\begin{figure}[h]
\begin{center}
\includegraphics[width=0.8\textwidth]{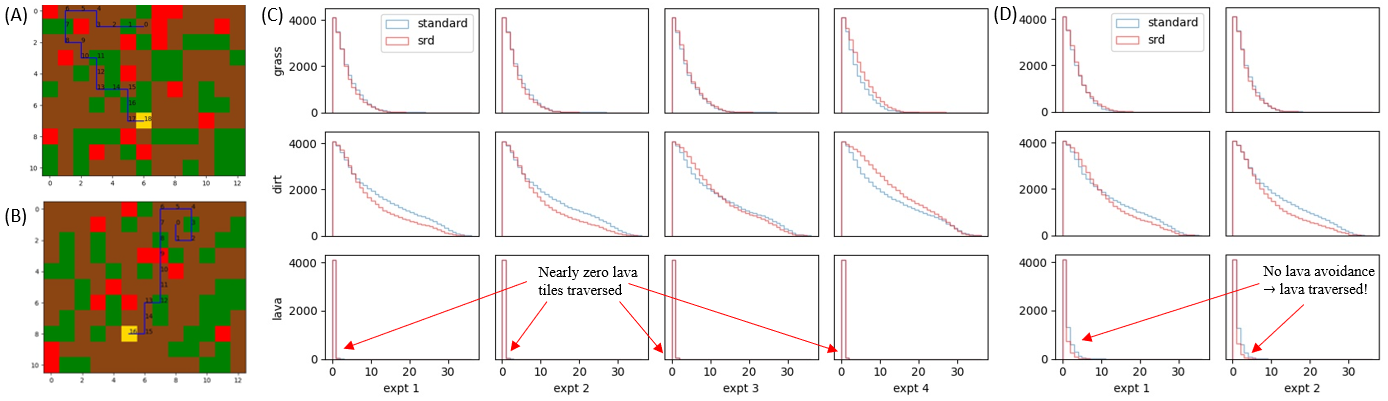}
\end{center}
\caption{(A) A sample trajectory from Lava A. (B) A sample trajectory from Lava NOAV A. A lava tile is traversed. (C) Cumulative histogram of the no. of tiles traversed by robot in Lava A. Almost zero lava tiles are traversed. (D) Similar to (B) but for Lava NOAV A, in which some lava tiles are traversed.}
\label{fig:lava}
\end{figure}

Comparison of results on different settings:
\begin{enumerate}[leftmargin=*,topsep=0pt]
\itemsep0em 
\item \textbf{Project A: standard design and SRD on maps without lava tiles}.  The fraction of grass tiles is \(0.3\). No robot will be spawned at the target immediately. Here, we use \(p_{dirt,grass}=0.2,-0.8\) chosen empirically. For training, we only run through 1 epoch, taking a very short time to complete (less than 0.5 hour per project without GPU). Fig. \ref{fig:no_lava}(B) shows the cumulative histogram of the number of tiles traversed with \(p_{target}=2\). 
\item \textbf{Compare A: model is more focused on the target}. This setting is similar to Project A, but we use a more extreme parameter setting, i.e. set \(p_{target}=10\). The local reward of the target tile is increased, and from table \ref{table:1}, we see that this generally increases the accuracies of both standard and SRD trained model (more likely to reach the target tile). The value seems to produce more unstable results as well, i.e. we empirically observe more failure modes.
\item \textbf{Project Lava A}. In this experiment, \(0.1\) of the tiles randomly assigned as lava and unknown avoidance \(u_a=2\). We also set \(p_{target}=10\) thus, similar to \textit{Compare A} we see higher accuracy but some failure modes as well. The model is designed to not recognize a lava tile, so a lava tile will activate the robot's Robot2NN \(w_{unknown}\) at lavas' positions. Fig. \ref{fig:lava}(A) is just an example of how robot successfully avoids lava, while fig. \ref{fig:lava}(C) lava cumulative histograms show that there are nearly zero lava tiles traversed.  
\item \textbf{Project Lava NOAV A}. Project Lava NOAV A is similar to Project Lava A, but we set \(u_a=0\), i.e. no unknown avoidance. The results are clear, we see in \ref{fig:lava}(D) that robot will traverse the lava tiles without much regards.
\end{enumerate}

\begin{figure}[h]
\begin{center}
\includegraphics[width=0.8\textwidth]{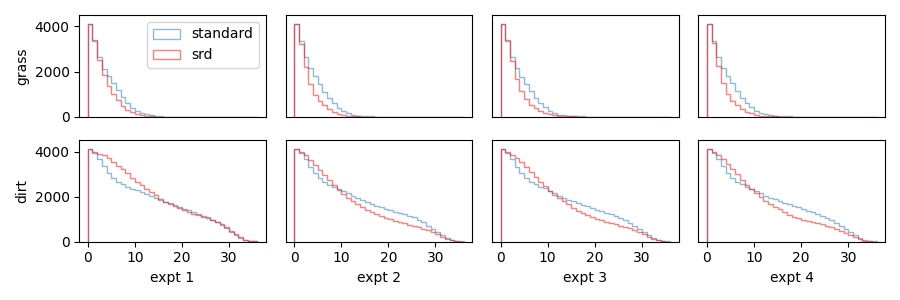}
\end{center}
\caption{Project B cumulative histograms.}
\label{fig:projBhist}
\end{figure}

\noindent Multiple trials of the above experiments have been conducted for reproducibility (project B, C, D etc). Each trial also consists of 4 experiments. Other results are in the supplementary materials (full version will be released later). The following are the names for the corresponding experiments. Repeat trials for Project A: project B, C and D. Histogram for project B can be seen at fig. \ref{fig:projBhist}. Repeat trials for Compare A: compare B, C and D. Repeat trials for Lava A: Lava B, C, D. Repeat trials for Lava NOAV A: Lava NOAV B, C, D.

\textbf{Robot2NN weights and preserved interpretability}. This model is very efficient because it consists of only 180 trainable parameters, as shown fully in fig. \ref{fig:weights}(B,C). As expected of relatively simple problems, there is no need for millions of parameters required to achieve high accuracy. High performance ~90\% accuracy is attained, given 10\% randomness is allowed. The weights of target deconv appear to have been trained towards higher positive values (redder). The center value remains the most prominent for all, thus preserving our interpretability. Looking into individual variations, fig. \ref{fig:weights}(B) shows the weights from Robot2NN model of project A expt 1 while fig. \ref{fig:weights}(C) from project Compare A expt 1. The difference in grass weights are apparent. 

We have seen from fig. \ref{fig:weights} that the trained models still have the general interpretable shape we initiated it with. While there is no theoretical proof, there may be an intuitive reason. Due to our interpretable design, the model starts off with a reasonable ability to solve the problem. This probably means  weights and biases already reside in a high dimensional parameter space somewhere around one of the local minima. This local minimum is special in the sense that it is more interpretable i.e. weights have recognizable shapes as we have initiated. As a result, a short training leads it nearer to that local minimum, hence the overall shape of the model remains similar to the initial shape and interpretable.

\textbf{(2) Why no conclusion should be drawn from this observation}? It seems that project A with smaller \(p_{target}\) results in relatively less preference for the grass tiles, which in turn leads to negative values (blue) for deconv for grass tiles. By comparison, project Compare A seems to have no negative values for grass tiles deconv. Unfortunately, the weights are shown only for demonstration; \textit{there is no definite conclusion that can be drawn}. This is because other experiments similar to project A also can result in all positive deconv weights with different patterns. They still yield high accuracy, thus possible variations within even this small set of parameters can still produce similar performance. Other results are shown in the appendix. Lava A project does not yield particularly distinct patterns. We see that even failure modes can yield weights profile that look similar to non-failure modes. Further investigations may be necessary.

\begin{figure}[h]
\begin{center}
\includegraphics[width=0.8\textwidth]{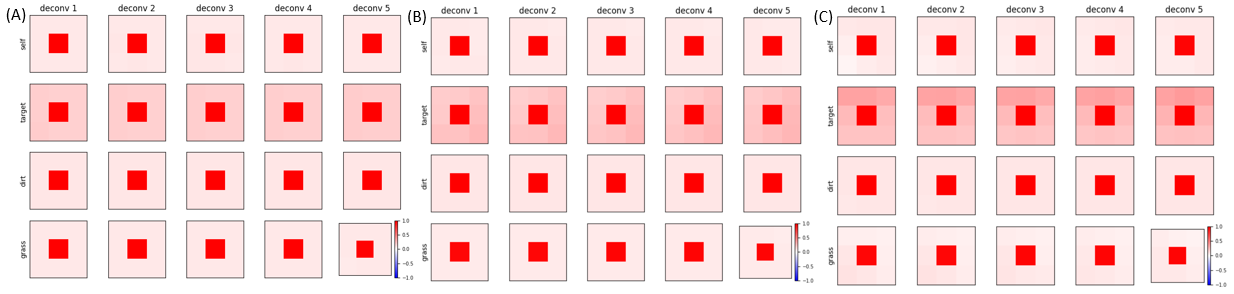}
\end{center}
\caption{Weights for (A) Project Lava A expt 1 (B) Project Lava NOAV A expt 1 (C) Lava A expt 4.}
\label{fig:weights_others}
\end{figure}

\begin{figure}[h]
\begin{center}
\includegraphics[width=0.8\textwidth]{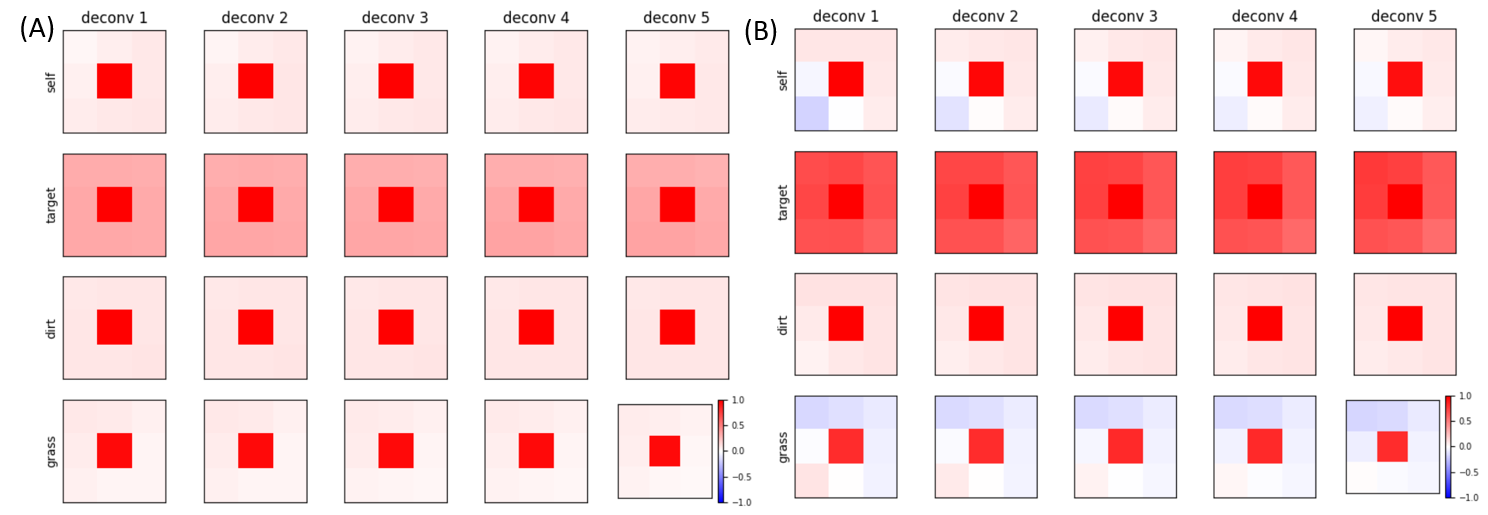}
\end{center}
\caption{Weights for failure modes (A) Project Compare A expt 2 (B) Project Compare A expt 4. (A) still shows standard-looking weights.}
\label{fig:weights_failure}
\end{figure}

\subsection*{MuJoCo with SRD}
We briefly described the application of SRD framework on half cheetah simulation on MuJoCo in the main text. Here, we will go through step by step process to arrive at the HalfCheetahSRD design (executed using \texttt{-{}-mode srd-model-design argument}); also, see \texttt{model\_half\_cheetah\_design\_stage.py}. The full set of commands used to execute our experiments can be found in \texttt{misc/commands\_mujoco.txt}.

\begin{figure}[h]
\begin{center}
\includegraphics[width=0.5\textwidth]{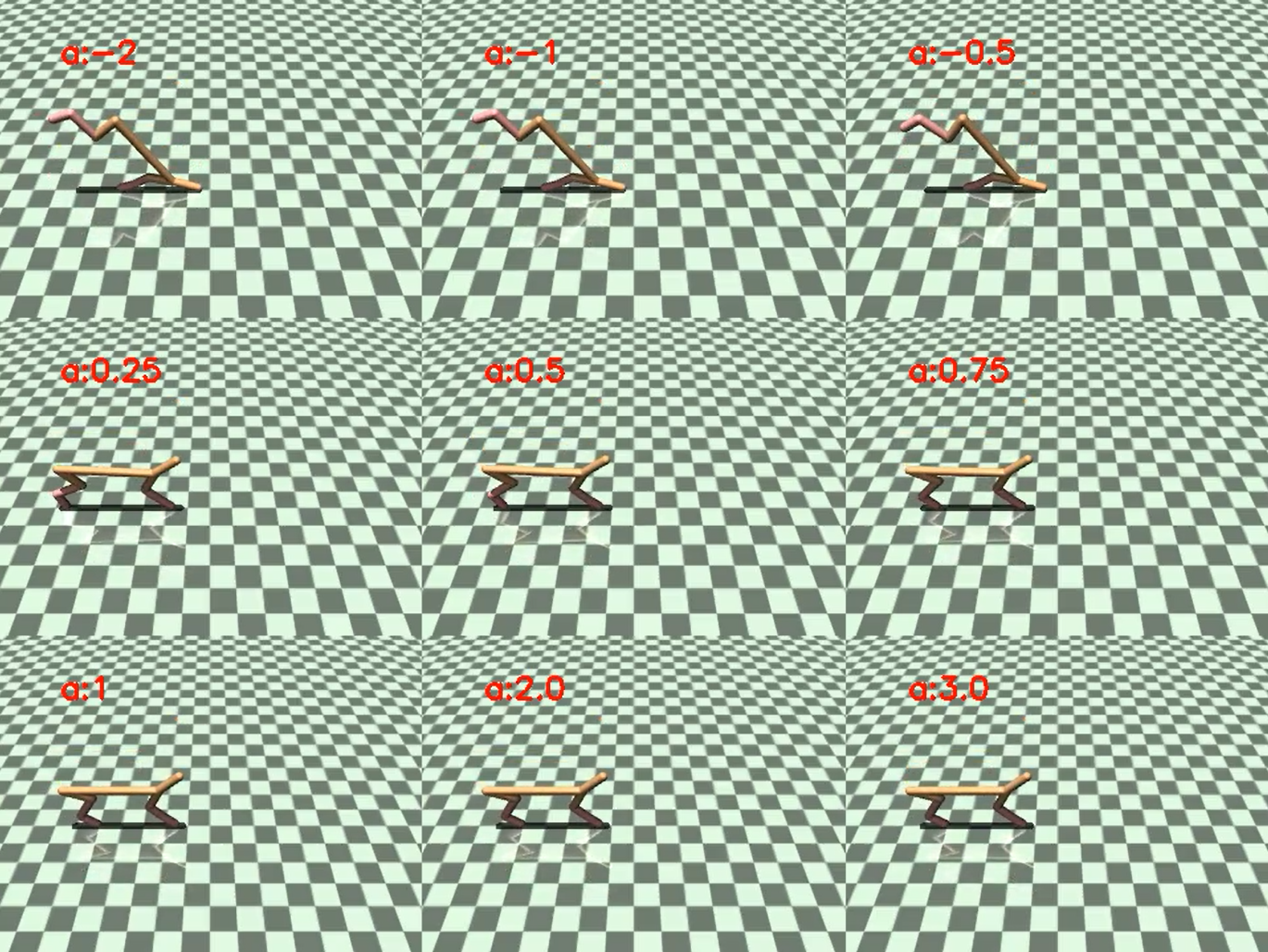}
\end{center}
\caption{Half cheetah's poses after the application of varying strengths of actuators (front and back thighs).}
\label{fig:poses}
\end{figure} 

\textbf{Stage 0: devtests}. To be able to design SRD properly, it is important to understand some fine details of the models and the platform used to simulate the model. We perform an initial testing to observe the agents' poses visually. Run the python command with \texttt{python mujoco\_entry.py -{}-mode devtests -{}-testtype vary\_control\_strength -{}-model half-cheetah}). Here, we arbitrarily choose a specific set of setup that we keep consistent throughout the experiment. For example, the framerate is set to 15 Hz (which does affect the time-step size) and \(interval=25\) (a setting for SRD model's momentum update). They are arbitrary; for our current experiments, all we need is for the agent to be able to stabilize and perform the running motion properly, noting that different settings might yield different results. 

At this point, we use some trial and errors to find a way to control actuators that will successfully make the half-cheetah run without using any neural network yet. Indeed, we found that the following works: (1) apply actuator with strengths \([0,0,0,-2s,0,0]\) for 25 time steps, (2) followed by \([-s,0,0,0,0,0]\)  for the next 25 time steps (3) repeat step (1) and (2) cyclically. The rationale is simple: step (1) is used to swing the front thigh forward and step (2) the back thigh. With this simple test movements, the agent is able to run forward. We will refer to this as the basis of the neural network we use in SRD design.

\textbf{Stage 1}. In this stage, our goal is only to observe the position coordinates of half cheetah. Different strengths of actuators are applied to the half cheetah and then it is allowed to stabilize. A short video will be available for readers to verify half cheetah's pose visually. The final poses are shown in fig. \ref{fig:poses}. To proceed with the model design, execute the python command to run stage 1: \texttt{python mujoco\_entry.py -{}-mode srd-model-design -{}-model half-cheetah -{}-stage 1}. We will then save these positions in \textit{init.params}, which we will use later. 

\begin{figure}[h]
\begin{center}
\includegraphics[width=0.9\textwidth]{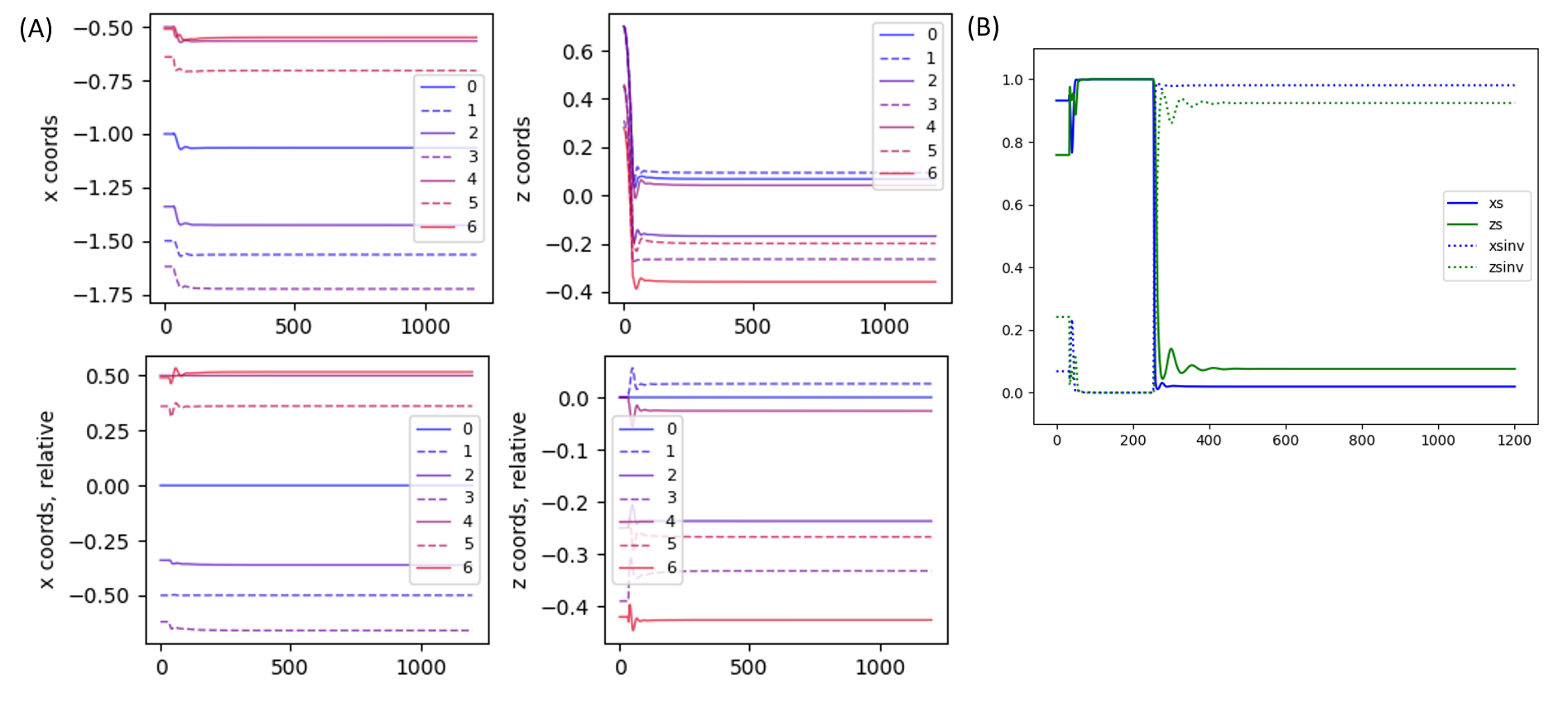}
\end{center}
\caption{(A) x and z position coordinates of the agent, half cheetah, in stage 1. Numbers 0 to 6 denote the 7 parts of half cheetah (B) The strength of activation of stable pose neurons \(xS,zS\) over time, and likewise of their inverses.}
\label{fig:stages12}
\end{figure}

In fig. \ref{fig:stages12}(A), numbers 0 to 6 denote the 7 parts of half cheetah as defined in the xml format (known as the MJCF model in the official MuJoCo documentation). For example, 0 corresponds to its torso, 1 to back thigh etc. The x and z position coordinates of these body parts relative to the torso's coordinates will be used as the input to the neural network that controls our SRD half-cheetah model, as in fig. \ref{fig:mujoco}(A). By observing the figure and the aforementioned short video, we verify that the agent is indeed initiated from a short distance above the ground, after which it will fall to the ground and stabilize on its front and rear legs.

\textbf{Stage 2}. In this stage, we consider how to convert the input (x and z coordinates) to a set of meaningful neuron activations. We start by considering neurons that respond to the stable standing pose from the previous stage. To achieve this, we use StablePoseNeuron, a custom pytorch module with parameter \(p\) and a forward propagation method that takes in input \(x\) and outputs \(\sigma_{sa}((x-p)^2)\). In fig. \ref{fig:mujoco}(A), xS and zS are both StablePoseNeuron, while \(xS_{inv}=1-xS\) and \(zS_{inv}=1-zS\). When the agent stays in a stable pose, \(xS,zS\) will activate strongly while their corresponding inverses \(xS_{inv}, zS_{inv}\) are not activated, and vice versa. This is shown in fig. \ref{fig:stages12}(B). In the scenario, the cheetah drops from a short height above the ground and stays in the equilibrium position until time step 250. From this point onward, the actuator of the agent's front thigh is activated, causing a forward swing of the front limb. This motion leads the agent away from its initial stable pose, and, as expected, the \(xS,zS\) neurons' signals drop off (and their inverses activate strongly). The goal of our stage 2 design has been achieved. Note: the command is the same as before, but with argument \texttt{-{}-stage 2}.

\textbf{Stage 3}. In this stage, we connect the stable pose neurons and their inverses to the actuators. At this point, we have a neural network structure similar to the blue dotted box in fig. \ref{fig:mujoco}(A), except without the BS neuron. Recall in stage 0 we swing back and front thigh alternately. Our aim here is to approximately replicate the set up, and then upgrade it with a neural network. Thus HalfCheetahSRD is born, in which the parameters of other actuators can be optimized in the SRD way as we have done before. More specifically, a fully connected layer connects the stable pose neurons to the actuators \(bt, bs, bf, ft, fs, ff\) as shown in fig. \ref{fig:mujoco}(A). The simulation is then run similar to the previous stage. We implement the \textit{momentum} function that ensures that the actuators apply their forces for 25 time steps before the next set of actuator's values are computed from the new, updated pose.

\begin{figure}[h]
\begin{center}
\includegraphics[width=0.9\textwidth]{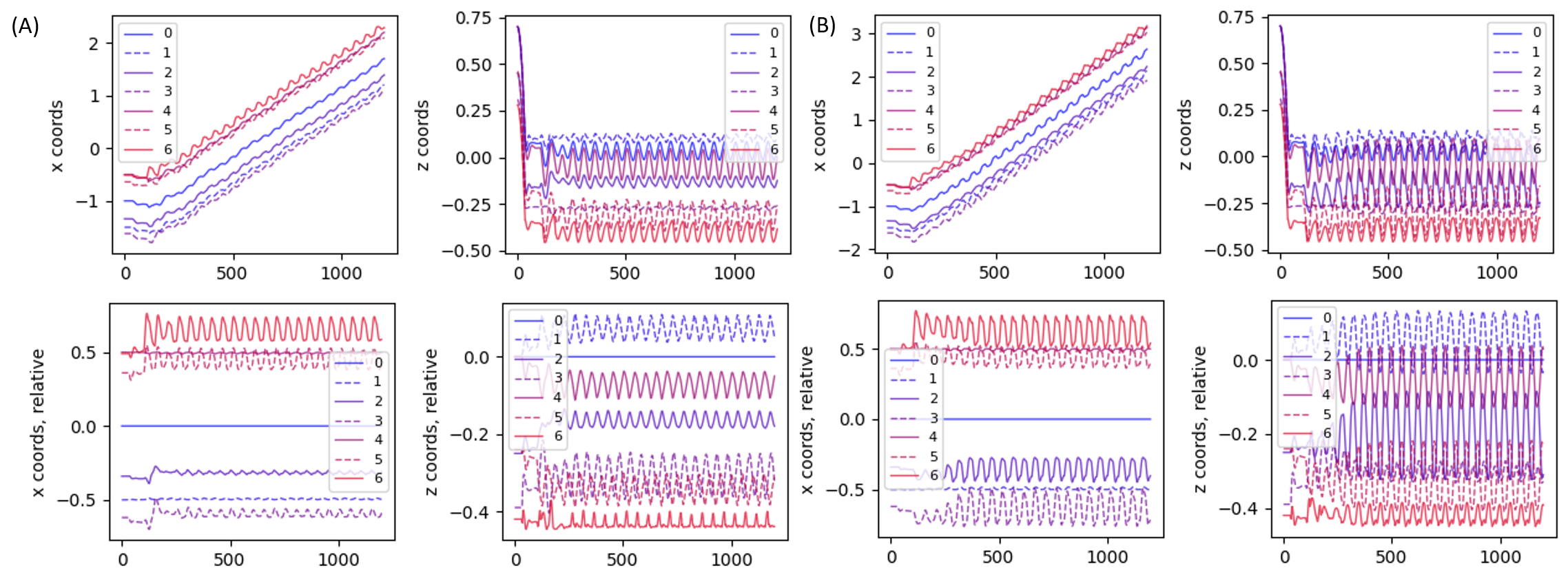}
\end{center}
\caption{(A) x and z position coordinates of the agent in stage 3. (B) x and z position coordinates of the agent in stage 4.}
\label{fig:stages34}
\end{figure}

The results are shown in fig. \ref{fig:stages34}(A). In essence, the plot of x coordinates shows that the agent moves forward successfully. The z coordinates drop from a height, as expected, and then oscillates regularly, indicating that the agent's body parts move in a regular cycle at a given level above the ground i.e. the cheetah does not trip or fly away etc.

Note: the command is the same as before, but with argument \texttt{-{}-stage 3}.

\textbf{Stage 4}. Stage 4 is similar to 3, except with the introduction of BS neuron with the modification of \texttt{propagate\_first\_layer} function. This neuron creates an additional variable used to vary the movement pose of the agent. More specifically, it allows the agent to vary how much its hind thigh swings throughout the motion. One such result with \(backswing=5\) is shown in fig. \ref{fig:stages34}(B).

\textbf{The main experiment}. The SRD is not yet complete without the PFC for SRD optimization. In this experiment, we keep the PFC simple, as shown in the green dotted box of fig. \ref{fig:mujoco}(A). As before, we want PFC to decide the correctness of the agent's action. The \textit{iN} neuron responds to the inhibitor, where the inhibitor takes the value of either 0 or 2. When \(inhibitor=0\), the agent will default to forward movement. But when \(inhibitor=2\), the agent is expected to stop moving. Without SRD optimization, the movements are plotted in fig. \ref{fig:mujoco}(C,D) bottom.

The PFC is designed such that the true \(T\) neuron activates more strongly when the actuator neuron \(Ac\) is activated while the \(iN\) neuron is not active. The false \(F\) neuron is approximately its reverse, penalizing motion when the inhibitor is active, i.e. when the controller wants the model to stop moving but the agent tries to move. The cross-entropy loss can be computed as before \(CEL(z,argmax(z))\), and the results are what we have discussed in the main text.

\end{document}